\documentclass[lettersize,journal]{IEEEtran}
\usepackage{url}
\usepackage{graphicx}
\usepackage{cite}

\usepackage{url}
\usepackage{subcaption}
\usepackage{amsmath}
\usepackage{amssymb}
\usepackage{amsfonts}
\usepackage{import}
\usepackage{breqn}

\def\eamp{{%
        \setbox0\hbox{/}%
        \rlap{\hbox to \wd0{\hss$\varepsilon$\hss}}\box0
}}

\usepackage{array, multirow, booktabs} 
\newcolumntype{L}[1]{>{\raggedright\let\newline\\\arraybackslash\hspace{0pt}}m{#1}}
\newcolumntype{C}[1]{>{\centering\let\newline\\\arraybackslash\hspace{0pt}}m{#1}}
\newcolumntype{R}[1]{>{\raggedleft\let\newline\\\arraybackslash\hspace{0pt}}m{#1}}

\begin{document}

\title{On the Effectiveness of Image Manipulation Detection in the Age of Social Media}

\author{Rosaura G.~VidalMata,~\IEEEmembership{Student Member,~IEEE}, Priscila Saboia, 
Daniel Moreira,~\IEEEmembership{Member,~IEEE}, Grant Jensen,
Jason Schlessman,
and~Walter J.~Scheirer,~\IEEEmembership{Senior Member,~IEEE}\\[2em] 
\thanks{Rosaura G.~VidalMata, Priscila Saboia, and Walter J.~Scheirer are with the Dept.~of Computer Science and Engineering, University of Notre Dame, IN.}
\thanks{Daniel Moreira is with the Dept.~of Computer Science, Loyola University Chicago, IL.}
\thanks{Grant Jensen is with Google LLC, Mountain View, CA.}
\thanks{Jason Schlessman is with Red Hat Research, Philadelphia, PA.}
\setcounter{figure}{0}   
\includegraphics[width=0.9\linewidth]{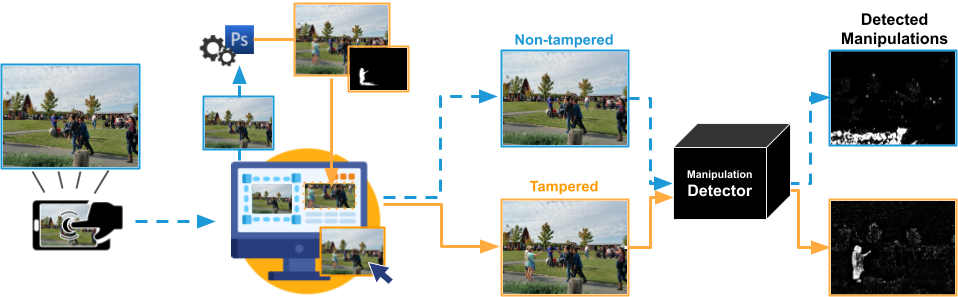}
\captionof{figure}{The image manipulation detection pipeline. In the right-most images presenting detection results, detected manipulated pixels are displayed in white and non-manipulated pixels in black. While manipulation detection approaches can localize the manipulated regions in a tampered image (bottom half of figure, continuous lines), the same approaches frequently misinterpret innocuous changes in non-tampered images as manipulations (top half of figure, dashed lines). The latter scenario is now widespread when analyzing images downloaded from the Internet, which almost always carry post-processing artifacts. }\vspace{-6mm}\label{fig:teaser}
}

\markboth{IEEE Transactions on Information Forensics and Security}%
{VidalMata \MakeLowercase{\textit{et al.}}: On the Effectiveness of Image Manipulation Detection in the Age of Social Media}


\maketitle

\begin{abstract}
Image manipulation detection algorithms designed to identify local anomalies often rely on the manipulated regions being ``sufficiently'' different from the rest of the non-tampered regions in the image. However, such anomalies might not be easily identifiable in high-quality manipulations, and their use is often based on the assumption that certain image phenomena are 
associated with the use of specific editing tools. This makes the task of manipulation detection hard in and of itself, with state-of-the-art detectors only being able to detect a limited number of manipulation types. More importantly, in cases where the anomaly assumption does not hold, the detection of false positives in otherwise non-manipulated images becomes a serious problem.
To understand the current state of manipulation detection, we present an in-depth analysis of  deep learning-based and learning-free methods, assessing their performance on different benchmark datasets containing tampered and non-tampered samples. We provide a comprehensive study of their suitability for detecting different manipulations as well as their robustness when presented with non-tampered data. 
Furthermore, we propose a novel deep learning-based pre-processing technique that accentuates the anomalies present in manipulated regions to make them more identifiable by a variety of manipulation detection methods.
To this end, we introduce 
an anomaly
enhancement loss that, when used with a residual architecture, improves the performance of different detection algorithms with a minimal introduction of false positives on the non-manipulated data.
Lastly, we introduce an open source manipulation detection toolkit comprising a number of standard detection algorithms.
\end{abstract}


\section{Introduction}


Widespread access to image editing applications has led to the explosive growth of image doctoring on the Internet~\cite{nymag}. While low-quality manipulations can often be easily identified through manual inspection, more advanced techniques frequently escape the human eye. The nature of these manipulations varies widely. For instance, an image can be manipulated without any ill intent for aesthetic or entertainment purposes. But the same manipulation techniques can be used to purposefully deceive viewers via the creation of malicious disinformation, which drives the need for tools to identify and localize such tampering.

Different techniques have been developed in the field of media forensics to classify images as forged or authentic as well as to localize and identify the tampered regions~\cite{li2009passive, ye2007detecting, ela_krawetz2007, Dirik2009ImageTD, Ferrara2012ImageFL, MAHDIAN20091497, lyu2014exposing, noi4_wagner2015, wu2019mantra}. Such approaches often look for artifacts introduced during the forgery process as different types of manipulations tend to introduce anomalies in the image's signal. These anomalies can come in the form of inconsistent noise~\cite{lyu2014exposing, MAHDIAN20091497, cozzolino2015splicebuster} and interpolation patterns~\cite{1511009, Dirik2009ImageTD, Ferrara2012ImageFL, gallagher2008image}, blocking artifacts due to re-compression~\cite{li2009passive, LIN20092492, IAKOVIDOU2018155, 8017428, bianchi2011improved, he2006detecting}, and other anomalies~\cite{johnson2006exposing, de2013exposing, kirchner2008fast}. Learning-free methods often look for these types of anomalies by analyzing the low-level image statistics whereas more recent deep learning-based approaches try to learn intrinsic image features that could be used to differentiate manipulated and non-manipulated data.  


With the rise of automated image post-processing (\textit{e.g.}, compression, image filtering) on social media platforms that emphasize image content, the entire endeavor of image manipulation detection, as is currently practiced, is now in question. The top half of Fig.~\ref{fig:teaser} shows an example of an input image with a splicing manipulation. 
While the inserted region appears to be visually coherent within the context of the image, its insertion introduced a set of anomalies that can be caught by pixel-level analysis tools. However, these tools are also sensitive to naturally occurring variations within the image, which can be incorrectly detected as tampering. This is shown in the bottom half of Fig.~\ref{fig:teaser}. Another problem exists in cases where the statistics of the spliced region are so similar to the ones extracted from the image's non-manipulated regions that the image might be able to pass undetected. As image manipulation tools get better, this is a growing concern. 

One way to address these problems is to refine the detection algorithm to make it sensitive enough to catch more subtle manipulations. Another way is to better identify features that are exclusively intrinsic to tampering (if they exist). However, given the wide variety of existing (and possible future) ways to manipulate an image, 
generating a method that can reliably identify their effects 
becomes challenging. Curiously, there has been little attention paid to the performance of detection algorithms when presented with pristine images, as they are often evaluated solely on data that has varying degrees of tampering. When used for the task of distinguishing between doctored and non-doctored images in a real-life scenario, one would expect the occurrence of false positives to be relatively low. However, in our investigations into this matter, we have determined that this is not always the case. 

In this article, we study different techniques for identifying local anomalies and review their performance on real-world forgeries as well as non-tampered data. In response to our experimental findings, we explore the use of image enhancement algorithms as a pre-processing step to improve the performance of the detection task such that  anomalies in images can be more easily identified by common manipulation detection approaches.


To facilitate the work of other researchers and to allow for the replication of the experiments that appear in this paper, we make a new image manipulation detection toolkit available for download, including trained models and training and evaluation source code in Python.

Our experiments and subsequent analysis are aimed at answering the following questions:
\begin{enumerate}  
    \item \textbf{How accurately can we differentiate between tampering-based anomalies and completely ordinary data?} This question relates to the situation in which the images do not contain any tampering but salient objects or naturally occurring features in the scene might be interpreted as so.   
    \item \textbf{Are certain local features better suited to prevent false positives?} While certain features have been proven to be a strong indicator of tampering, naturally occurring changes in the image (not related to tampering) can be mistaken for manipulations. Are certain features more prone to this?
    \item \textbf{How does the preservation of pixel-level anomalies and deep-generated anomalies affect detection?} Here we seek to assess the impact of a pre-processing step to preserve/increase the presence of certain anomaly cues used by manipulation detection approaches. Is this pre-processing better for certain scenarios and worse for others?
\end{enumerate}

In the rest of this article, we first review the general background research related to manipulation detection in Sec.~\ref{sec:related}. We then go on to provide detailed descriptions of the approaches most commonly used in practice in Sec.~\ref{sec:algorithms}. These approaches form the basis for the new forensics toolkit we are releasing with this paper called Python-based Image Forgery Detection (pyIFD) toolkit. For a common way to improve the performance of manipulation detectors, we introduce the idea of an Anomaly Enhancement Network in Sec.~\ref{sec:aen}. To gain a sense of how effective manipulation detectors are on data from the contemporary Internet, we introduce a large-scale experimental study in Sec.~\ref{sec:experiments} and discuss our findings in Sec.~\ref{sec:results}. Finally, we end with some thoughts on where manipulation detection research needs to go in the future.

\section{Related Work}
\label{sec:related}

Research in media forensics has received growing attention from academia, industry, and government agencies. Representative of this has been initiatives such as the Media Forensics (MediFor) program~\cite{Medifor}, launched by the Defense Advanced Research Projects Agency (DARPA) of the U.S. Department of Defense to push for the development of image and video integrity assessment, as well as the creation of media forensics datasets and challenges~\cite{mfc18_challenge, OpenMFC}. A key piece of that effort was the evaluation of existing manipulation detection approaches and the creation of new ones responding to newly identified weaknesses.  

Despite the growing number of manipulation detection approaches, there have been few studies providing a comprehensive analysis of the state of the field~\cite{zampoglou2017large, 6301704}, with most of the work focusing on methods designed to identify splice or copy-move forgeries. Zampoglou et al.~\cite{zampoglou2017large} study traditional forgery detection methods, providing further analysis on their strengths and limitations, taking into account that data on the web would more often than not undergo multiple rounds of re-compression and scaling.
They analyze 
the methods' performance when faced with data with varying levels of compression as well as web-based images.
Similarly, Christlein et al.~\cite{6301704} analyze the performance of various forgery detection algorithms and their robustness to different post-processing scenarios such as the introduction of Gaussian noise, re-scaling, rotation, and JPEG compression. Nevertheless, both of these works limit their analysis to studying the detectors' performance when presented with images containing some degree of tampering. While this is a valuable step in a pipeline where the algorithms are presented with images that have been identified as forgeries, there is no study on their robustness when presented with pristine data, which is something we address in this paper.


There have been many methods proposed to detect and localize different types of manipulations. Early approaches were designed to localize specific types of forgeries through an analysis of low-level tampering artifacts such as JPEG compression~\cite{bianchi2011improved, li2009passive, ye2007detecting, ela_krawetz2007}, anomalies in the Color Filter Array (CFA) patterns~\cite{Dirik2009ImageTD,Ferrara2012ImageFL,goljan2015cfa}, noise variance inconsistencies~\cite{MAHDIAN20091497, lyu2014exposing, noi4_wagner2015}, and illumination-based methods~\cite{carvalho2015exposing, de2013exposing}. While generally effective, most of these techniques depend strongly on scenario-dependent assumptions. 

JPEG compression-based techniques work under the assumption that all the regions in a natural image should have undergone the same amount of compression, while spliced regions would be subject to a different compression regime than the background image they are spliced into. Similarly, CFA localization methods assume that since the CFA pattern and interpolation technique used to create a digital image varies from sensor to sensor, there would be a sufficient difference between the pattern in the manipulated and non-manipulated regions in an image to detect and localize doctored regions. However, this might not hold true in cases where the tampered regions have a similar CFA pattern (\textit{e.g.,} in cases where both the spliced and background regions were created using the same type of sensor) or when the full image has been resized (as this would destroy the original CFA pattern of the non-tampered regions). 

Noise analysis methods work under the premise that natural images have a somewhat uniform global noise pattern that is distorted by local manipulations. As such, the difference between the global and local noise characteristics could be used as an indicator of tampering. This falls through when post-processing techniques like filtering are applied after splicing as they would erode any of the anomalies these algorithms depend on. Illumination-based methods are dependent on uniform global light patterns as they rely on inconsistencies in light source direction or perspective mappings between the background and spliced regions. While often more reliable than the other methods, their performance is dependent on the size of the tampered regions, as smaller regions leave the global illumination patterns mostly unaltered.

Deep learning-based techniques have now become commonplace for manipulation detection, with early approaches patterned after the traditional methods to exploit artifacts left behind by the manipulation method, such as JPEG compression artifacts~\cite{chen2008determining, 8017428, he2006detecting, park2018double}, noise residuals~\cite{zhao2013image, he2012digital, li2016identification}, or boundary artifacts~\cite{salloum2018image, bappy2017exploiting}. Most of the current approaches focus on a particular source of tampering evidence or on a specific manipulation type like splicing~\cite{chen2017image, cozzolino2015splicebuster, huh2018fighting}, copy-move~\cite{7823911, wen2016coverage, 7154457, wu2018busternet}, or in-painting~\cite{zhu2018deep, 7026073}. They obtain strong performance for the forgery method they were trained on, but with limited applicability on unseen forgery methods. To address this, some approaches aim at learning features related to multiple local anomaly types~\cite{wu2019mantra}, while others treat forgery localization as a single-class anomaly detection problem~\cite{d2017autoencoder,cozzolino2016single,cozzolino2018forensictransfer}.

Following this trend, during the training of our proposed pre-processing network, we treat any manipulation as an anomaly without differentiating between the specific kinds of tampering methods. Since our method's learning is done independently from each of the detection approaches tested, the features learned and enhanced by our network are general enough to capture and emphasize multiple types of pixel-level anomalies used by common detection approaches. To our knowledge, this paper is the first one to introduce a pre-processing-based approach designed to improve the performance of existing manipulation detection approaches.

\section{Manipulation Detection Algorithms}\label{sec:algorithms}
\subsection{pyIFD: A New Manipulation Detection Toolkit Comprised of Familiar Algorithms}
In this section we provide a more detailed review of the learning-free local anomaly detection strategies most commonly used in practice. These approaches remain viable even in the era of deep learning, and in some cases are superior to the 
machine learning-based ones
because they are not dependent on training data tied to specific settings. Reference implementations for these approaches that can be incorporated into a large-scale evaluation framework have been hard to come by, with many implemented in legacy Matlab or Java code~\cite{MKLab}. To support our own study, and as a service to the media forensics community, we created the Python-based Image Forgery Detection (pyIFD) toolkit\footnote{Available at \url{https://github.com/EldritchJS/pyIFD}.}.
The pyIFD library is open source and written in Python to provide a cross-platform package to researchers and practitioners. It is available via PyPi to facilitate quick installation of the included algorithms. These algorithms provide strategies to estimate the anomalies left behind by different manipulation types, such as the introduction of compression artifacts, discrepancies in the image's interpolation or noise patterns, or the extraction of 
features that encode the differences between natural and anomalous data.

\textbf{JPEG Compression Artifacts.} JPEG compression takes advantage of the fact that the high-frequency information in an image is not easily perceived by the human eye. As such, it discards this type of information when encoding the image data. When performing JPEG compression the image is split into 8x8 pixel groups, each of which is further encoded separately using its own Discrete Cosine Transform (DCT). Each of these groups can be replicated by 64 cosine waves. Thus, the DCT is used to calculate the contribution of each of the waves, which often results in the low-frequency ones having a larger effect than the high-frequency ones. During quantization, the high-frequency data on each of the blocks is removed by dividing each of the coefficients by their corresponding quantization value (specified by the compressor's quantization table and the JPEG quality setting) and rounding to the nearest integer. The output is then serialized to provide encoding. 

This process introduces horizontal and vertical breaks into the image known as blocking artifacts. In this work, we consider three different approaches that look at different inconsistencies caused by these types of artifacts:

\begin{itemize} 
    \item \textbf{BLK~\cite{li2009passive}}: This approach is based on the observation that one of the effects of 
    content cloning
    operations is the creation of a mismatch between the 
    added
    and original content in the blocking artifact grid of an image. Taking this into account, BLK models the blocking artifact grid extracted out of an image's luminance component and uses the inconsistencies in the grid to identify manipulated regions. However, it is important to note that the variations in the grid might also be a product of the image content, which makes these methods sensitive to false positives due to non-JPEG features. 
    \item \textbf{DCT~\cite{ye2007detecting}}: Since different JPEG-compressors use different quantization tables, the regions within a tampered image might present different types of compression artifacts if they were compressed using different tables. This approach is based on a quantization table estimation, which is used to calculate a blocking artifact measure to identify discrepancies in the local JPEG quantization history of each block.
    \item \textbf{ELA~\cite{ela_krawetz2007}} (Error Level Analysis): Given that JPEG is a lossy format, as an image undergoes consecutive rounds of compression, it loses more and more of its high-frequency content up to the point where further re-compressions cause no change (local minima for error). ELA works under the assumption that manipulations added to an image would have undergone fewer re-compressions than the rest of the image. As such, the difference between these ``newer'' regions across consecutive rounds of compression would be higher than the ones for regions that had already reached their local minima. 
\end{itemize}

\textbf{Discrepancies in Color Filter Array (CFA) Interpolation Patterns.} Digital images are often captured using a single sensor through which the light is filtered using a CFA. As this would produce a single value per pixel, the images are then transformed into three channels using some interpolation algorithm. However, as the CFA and the interpolation algorithm used to generate an image might not only vary from sensor to sensor but be sensitive to geometric transformations, content splicing manipulations could generate discrepancies in the CFA pattern of an image. Here we consider the following CFA-based methods:

\begin{itemize} 
    \item \textbf{CFA1~\cite{Dirik2009ImageTD}}: This approach uses two features to localize tampering in an image. The first is based on estimating the CFA pattern of the source sensor. Discrepancies between the estimated and observed values are then used to detect local tampering. The second uses the variances between the noise of interpolated and natural pixels to determine the probability of the pixel values having been disrupted by tampering.
    \item \textbf{CFA2~\cite{Ferrara2012ImageFL}}: This approach also measures the presence of local interpolation artifacts between the interpolated and observed values. However, it does so by estimating the tampering probability in 2x2 blocks, which allows for a finer-grained localization of tampering. 
\end{itemize}
It is important to note that both of these approaches are sensitive to strong JPEG re-compression and resizing, as these type of operations tend to distort/suppress CFA artifacts.

\textbf{Noise Variance Inconsistencies.} 
Noise can be introduced into an image during its capture, compression, transmission, and other post-processing operations. While noise deteriorates the image quality, it tends to be uniformly distributed across the entire image. It has been observed that local manipulations may introduce inconsistencies in the image's global noise pattern, which makes that pattern a good candidate for observation when looking for tampered regions in an image. Several approaches have been proposed to analyze the noise patterns in an image to detect manipulations such as splicing, airbrushing, and warping. In this work we consider the following methods:  

\begin{itemize} 
    \item \textbf{NOI1~\cite{MAHDIAN20091497}}: This approach analyzes the local noise level inconsistencies to detect and segment regions corrupted by local noise (the approach assumes a white Gaussian noise with a variance that can vary spatially). For this, the local noise is estimated by tiling the high pass diagonal wavelet coefficients with non-overlapping blocks (the blocks are assumed to be smaller than the size of the corrupted regions), with the standard deviation for each block used as a homogeneity condition to segment the image into sub-regions. This method is sensitive to variations in the local frequency spectrum, which makes it prone to generating false positives. Additionally, it is not able to find corrupted regions when the noise degradations are very small, which results in coarse localizations of the tampered areas.
    \item \textbf{NOI2}~\cite{lyu2014exposing}: This approach is based on the observation that natural images in band-pass domains often display a statistical regularity: their kurtosis values tend to be close to a positive constant (projection kurtosis concentration). Taking this into account, NOI2 formulates a blind global noise estimation algorithm based on this property and extends it to estimate locally varying noise levels. The local noise estimation is then used to detect any possible spliced regions within the image. The estimations obtained from this approach are approximations of the local noise variances, which for tampering detection purposes are considered to be a sufficient 
    indicator
    of splicing. However, this might not be the case for images with a large number of small-scale textures, which could be categorized as false positives.
    \item \textbf{NOI4}~\cite{noi4_wagner2015}: Since manipulations on an image often tend to leave visible traces in the image's noise map, this approach uses the noise variance inconsistencies between an image and its de-noised counterpart as a measure for tampering. For this, median filtering is performed on an image to obtain the difference between the original and median-filtered image: the median filter residual. The residual is then used as the image's forensic fingerprint. 
\end{itemize}

\subsection{Deep Learning-Based Methods} 
In addition to the methods included in our manipulation detection toolkit, we also consider two state-of-the-art deep learning-based manipulation detection approaches in this work. These types of detectors rely on learning features that describe possible manipulations. While some approaches aim to improve the detection of known artifacts, such as the ones addressed by the previously discussed approaches, others look for specific types of manipulations that are not explicitly defined or try to extract compact features that can be analyzed to identify manipulations at the patch level.

\textbf{Mantra-Net~\cite{wu2019mantra}} is a Manipulation Tracing Network for the localization of image forgeries. One of the key aspects of Mantra-Net is its ability to detect multiple types of manipulations. The Mantra-Net architecture is composed of two main blocks: (1) an Image Manipulation Trace Feature Extractor and (2) a Local Anomaly Detection Network. 

Mantra-Net uses VGG~\cite{VGG:2014} as the backbone for the feature extractor, which is trained to classify a hierarchy of image manipulations encompassing over 385 types of fine-grained manipulations. After training the feature extractor, the decision block is discarded and the  extracted features are used as the input for the Local Anomaly Detection Network. This Local Anomaly Detection Network uses a module based on Long Short-Term Memory (LSTM), which is trained jointly with the pre-trained feature extractor. Given the feature map, the objective is to identify the dominant feature of the image and deem any feature that is sufficiently different from this dominant feature as anomalous. 

We use the pre-trained weights provided in the Mantra-Net reference implementation for our evaluation. To account for the large amount of computation that the network performs, we crop the input images into smaller blocks (each block has at most 160K pixels) and process them individually before joining them together to generate the full manipulation mask.

\textbf{RGBN~\cite{zhou2018learning}} is a two-stream network based upon the Faster R-CNN~\cite{ren2015faster} object detection network.
Instead of detecting objects, this network is trained to localize tampered regions within a host image, covering diverse manipulation techniques such as splicing, copy-move, and removal.
While one of the streams is focused on the RGB values of the analyzed images, the other parallel one is focused on the images' generation and capture noise (N).
The underlying assumption is that the RGB stream specializes on strong-contrast manipulation artifacts (such as unnatural boundaries), while the N stream specializes on anomalies caused by disturbing the global image noise.
The two streams are then combined through a bilinear pooling layer (hence the herein adopted RGBN name).

We use the implementation provided by the authors and follow their recipe to train RGBN, using the same dataset (COCO~\cite{lin2014microsoft}) and synthetic manipulation generation processes~\cite{rgbn}.

\begin{figure}[t]
\centering
\includegraphics[width=\linewidth]{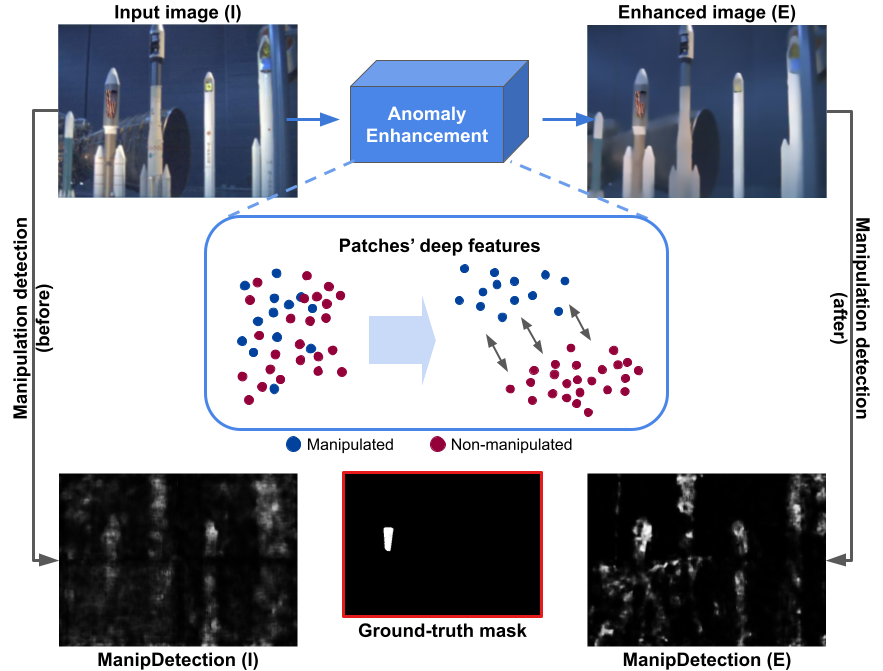}
\caption{Anomaly Enhancement Pipeline. The proposed Anomaly Enhancement Network (AEN) reconstructs an input image $I$ such that the anomalies between the tampered regions and their neighboring pixels are better emphasized in the image's enhanced version $E$.
They can then be more easily spotted by a manipulation detection algorithm.}
\label{fig:anomalyenhancement}
\end{figure}

\section{Proposed Anomaly Enhancement Network} \label{sec:aen}
With widespread access to image editing tools, the volume of doctored images present in social media has increased dramatically. In addition to this, with the increased quality of these manipulations, even humans struggle with spotting any traces left behind by tampering. One way to address these problems is to refine the detection algorithms to make them sensitive enough to catch more subtle manipulations. Another way is to identify better features exclusively intrinsic to tampering (if they exist). However, identifying these features is a challenging task given the wide variety of existing (and possible future) ways to manipulate an image. Curiously, there has been little attention paid to 
finding and learning to enhance image features 
as a pre-processing step to improve the performance of 
preexisting manipulation detectors.

\begin{figure*}[t]
  \centering
  \begin{tabular}{c c}
    \includegraphics[width=0.45\textwidth]{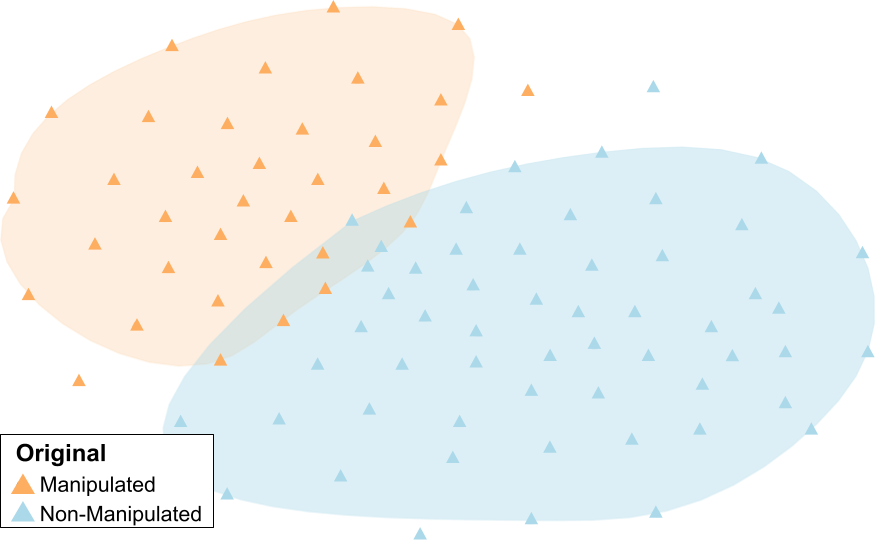} &  \includegraphics[width=0.45\textwidth]{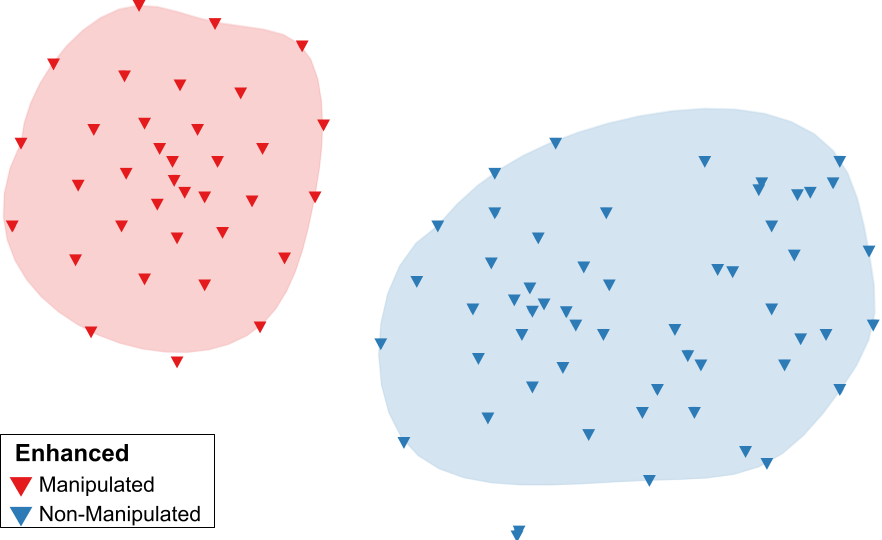} \\
    \small (a) Original & \small (b) Enhanced
  \end{tabular}

  \caption{Visualization using t-SNE~\cite{wattenberg2016how} for non-linear dimensionality reduction of the embedded features of manipulated and non-manipulated patches extracted from one of the images from the MFC18 Dev Dataset (see Sec.~\ref{sec:dataset}).
   Our proposed AEN method is able to increase the distance between the manipulated and non-manipulated features as well as promote tighter clusters for patches belonging to the same class.} \label{fig:ft_embedding_vis}
\end{figure*}

As such, we study the suitability of one such method, and propose a novel Anomaly Enhancement Network (AEN) to be introduced as a pre-processing step in a manipulation detection pipeline (see Fig.~\ref{fig:anomalyenhancement}). 
We base our work on the observation that in embedding spaces, image patches containing manipulated pixels often show a different distribution than the rest of the (non-manipulated) pixels in the analysed image (see Fig.~\ref{fig:ft_embedding_vis}a). 
However, the separation between the distributions of manipulated and non-manipulated data might not be as prominent depending on the type, area, and context of the manipulation.

Motivated by this, our pre-processing tool seeks to emphasize the subtle local differences between natural and tampered pixels in an embedding space to both increase the distance between their distributions and make their distributions more individually cohesive (see Fig.~\ref{fig:ft_embedding_vis}b).
We propose to do this through a learned data-driven image content reconstruction, which preserves the original pixel-level discrepancies that manipulation detectors look for, while introducing further disparities in the tampered regions, as a way to facilitate the detection of more subtle manipulations. 





Consequently, when reconstructing each region in the image, we consider three main objectives: 
\begin{enumerate}
    \item \textbf{Content preservation}: Given the different sensitivities of the manipulation detection approaches, it is important to preserve as much of the pixel-level information in the reconstructed regions in order to avoid eroding any of the anomalies that they might be tuned to find.
    \item \textbf{Intra-class cohesion}: The loss of information and the possible introduction of noise and blurring artifacts during reconstruction has the potential of making non-manipulated regions similar to their tampered counterparts. To alleviate this effect we seek to increase the similarity between regions containing the same type of data (manipulated or non-manipulated) in an effort at preserving the features intrinsic to each of our classes.
    \item \textbf{Anomaly enhancement:} Regardless of the content of each reconstructed region, we seek to emphasize the differences between manipulated and non-manipulated pixels within the image to facilitate the identification of tampered regions.
\end{enumerate}



\subsection{Loss Function}
Consider our three main objectives.
While we can optimize the content restoration using a traditional pixel-level loss (\textit{e.g.}, $L_{1}$), in order for our network to be able to emphasize the anomalies caused by tampering while ensuring the enhancement preserves the characteristics of each region, it must also learn the differences between manipulated and non-manipulated data. For this, we present a reconstruction network with a series of triplets, 
each containing (1) an anchor patch $a$ of a given class $c \: \epsilon \: \{c_{1}, c_{2}\}$ (manipulated, or non-manipulated), (2) a positive patch $p$ belonging to the same class as $a$, and (3) a negative patch $n$, belonging to the opposite class as $a$ (\textit{i.e.}, if $a$ belongs to the manipulated class, $n$ would be a non-manipulated patch, and vice versa).

We then seek to produce an enhanced version of our anchor patch, $\hat{a}$, that closely resembles the input anchor $a$ in the pixel space and other members of the anchor's class ($p$) in a deep embedding space. To ensure the regions in the reconstructed image highlight the tampered regions, the reconstructed patch $\hat{a}$ must also diverge from patches belonging to the contrasting class in an embedding space.

The goal of training can then be summarized by minimizing the following loss function: 
\begin{equation}
\begin{split}
L(a, \hat{a}, p, n) = w_{0} D(a, \hat{a}) + w_{1} D(f(\hat{a}), f(p)) \\- w_{2} D(f(\hat{a}), f(n)).
\end{split}\label{eq:loss}
\end{equation}
The loss 
consists of three main components. The first one corresponds to the reconstruction fidelity, measuring the distance between the reconstructed patch $\hat{a}$ and the input $a$ in a pixel-level domain. The second is the intra-class similarity, which ensures elements from the same class are close to each other in a deep embedding domain. The last one is the anomaly enhancement factor, which seeks to increase the distance between the features extracted from manipulated and non-manipulated patches.

Within the loss, $D(x,y)$ is a distance function, $f(x)$ represents the deep features extracted from a patch $x$ using a 
CNN pre-trained on ImageNet~\cite{ILSVRC15}, and $w_{i}$ corresponds to the weight for each component of the loss. 
Here we used the Symmetric Mean Absolute Percentage Error (SMAPE)~\cite{MAKRIDAKIS1993527} as the distance metric $D$ to normalize the distances across each of the terms of the loss and to constrain it between $0$ and $1$. 

We provide an analysis of the impact of each of the three components of our loss and evaluate the performance of our network given three weight configurations, each favoring one of the main objectives of the proposed loss in the Supplementary Material.

\subsection{Network Architecture.}
Our approach is designed to re-construct an input image $I$ while highlighting the differences between any manipulated regions and their non-manipulated vicinities. Given the success of residual neural networks in the area of image reconstruction~\cite{zhang2018image, Anwar_2019_ICCV, Wang_2018_ECCV_Workshops} we use one of such architectures as the backbone for our method. Our anomaly enhancement network is inspired by the Residual Channel Attention Network (RCAN) proposed by Zhang et al.~\cite{zhang2018image} 
The network 
uses a single convolutional layer for shallow feature extraction, followed by five residual attention groups (RG) with a long connection between the shallow feature layer and the convolved output of the last residual group. Each residual group contains 10 residual channel attention blocks (RCAB) connected by short skip connections. These channel attention blocks employ a channel attention mechanism that models the inter-dependencies across the different feature channels to focus on the ones providing more informative features. The output patch is then reconstructed out of the aggregated output of the last residual group and the shallow feature layer. We take advantage of the long skip connection to preserve low-frequency features while the rest of the network focuses on the more informative high-frequency channel-wise features in the input patch.
This framework is then trained using the above loss (Eq.~\ref{eq:loss}) until convergence. The model is saved at the epoch with minimal training loss.

\section{Experimental Setup}
\label{sec:experiments}

In this section we detail the experimental framework that will let us evaluate the effectiveness of both learning-free and deep learning-based image manipulation detectors. Importantly, we emphasize performance on known clean data that has not been manipulated in a malicious way. This is a key aspect that has been missing from previous evaluations, including the DARPA MediFor program. Because it has been missing, many prior studies overemphasize the performance of the various detectors.

\subsection{Datasets}\label{sec:dataset}
We use two datasets from the Media Forensics 2018 Challenge (MFC18)~\cite{8638296} to evaluate the detection of manipulated regions. Additionally, we analyzed our method's performance when presented with pristine images extracted from ImageNet (Imagenette~\cite{imagenette}) and the UG$^2$ Dataset~\cite{8354281}.

\subsubsection{The MFC18 Datasets} these datasets consist of images manipulated by hand by a group of expert manipulators using different media editing tools, showcasing a variety of local (\textit{e.g.}, splicing, in-painting) and global (\textit{e.g.}, blur, sharpening) manipulation operations. For this work, we used the \textit{MFC18 Dev} and \textit{MFC18 Eval} datasets,
which contain manipulated images as well as their reference ground-truth masks for localization, where each bit indicates the manipulated region(s) of a distinct manipulation operation step. It is important to note that the ground-truth masks used in this work do not reflect global manipulations (\textit{e.g.}, blur) affecting the entire image.   

The proposed pre-processing step for anomaly enhancement was trained using triplets of manipulated and non-manipulated patches extracted from a subset of the MFC18 Dev Dataset. 
After training, we processed the images from both the MFC18 Dev and Eval Datasets (757 and 236 images respectively) and fed them to a wide variety of manipulation detection approaches to evaluate their performance (see Sec.~\ref{sec:algorithms} for a description of the detection algorithms used as well as their performance).  

\subsubsection{UG$^2$ Glider} this is a subset of 2,778 images extracted from non-tampered frames belonging to the UG$^{2}$ Dataset~\cite{8354281}. The frames were extracted out of 38 unique videos and showcase 15 distinctive object classes. We use this dataset as the entirety of its data does not contain any type of manipulation and was subjected to minimal rounds of compression.

\subsubsection{Imagenette} this is a subset of images extracted from the ImageNet dataset belonging to 10 distinctive object classes. We are using the validation partition (consisting of 3,925 images) for our work here. While we consider this dataset to contain data that was not changed in a malicious way, as it was collected from popular sites on the Internet, some of its images might have been subjected to global manipulations to either improve the aesthetics of the images or to add a watermark to the pictures. Nevertheless, given the nature of these manipulations and their relevance to a real-world manipulation detection task, we consider them to be ``clean'' data and evaluate them as such.

\subsection{Evaluation Metrics}
We evaluate the performance of each of the discussed algorithms following the metrics defined by the MFC18 Challenge~\cite{mfc18_challenge}.
All of the following metrics were computed through the Mediscore tool~\cite{mediscore_nist_2021}, which was provided by the authors of the dataset. 

Most metrics depend on having binarized values for the predicted manipulation masks, whereas the localization algorithms more often than not generate a gray-scale localization mask, where the value of each pixel ranges from 0 to 255. While it is possible to binarize the predicted masks by either specifying a binarization threshold $\theta$, the selection of such a threshold then impacts the performance and sensitivity of such metrics. Taking this into account, we use two metrics that do not depend on binary masks: Area Under the Curve (AUC) and Grayscale Weighted L1 Loss (GWL1).
In summary, the higher the AUC, the better the assessed manipulation detection solution is.
Contrary to the AUC, the lower the GWL1 value, the better the solution.
A complete description of both metrics is provided in the Supplementary Material that accompanies this manuscript.

Since the Mediscore tool does not support the analysis of non-manipulated images, when evaluating the performance of non-manipulated data (Imagenette and UG$^2$ datasets), we mark a small arbitrary region (10x10 pixels) in each image as manipulated and use it as ground truth.
This allows us to still use Mediscore to compare and rank the different solutions when they are presented with non-manipulated images, at the cost of a small penalty for the 10x10 regions, equally applied to all the images and methods. 
Given the small area of this fictional tampering, any outputs passing as true positives 
have little impact on the final loss score.

\section{Results}\label{sec:results}

In this section, we provide our analysis for manipulation detection performance, non-manipulated data performance, and the use of anomaly enhancement as a pre-processing step. 

\subsection{Manipulation Detection Performance}
As discussed in Sec.~\ref{sec:algorithms}, the features used by the evaluated algorithms can be roughly divided into four categories: (i) Compression Artifacts (BLK~\cite{li2009passive}, DCT~\cite{ye2007detecting}, ELA~\cite{ela_krawetz2007}), (ii) Color Filter Array Discrepancies (CFA1~\cite{Dirik2009ImageTD}, CFA2~\cite{Ferrara2012ImageFL}), (iii) Noise Variance (NOI1~\cite{MAHDIAN20091497}, NOI2~\cite{lyu2014exposing}, NOI4~\cite{noi4_wagner2015}), and (iv) Deep Learning Features (MantraNet~\cite{wu2019mantra} and RGBN~\cite{zhou2018learning}).
We can see in Table~\ref{tab:all_base} the GWL1 and AUC for each of the tampering localization methods belonging to these categories, for both the MFC18 Dev and MFC18 Eval datasets (in the first and second column groups, respectively). 

\begin{table*}[!t]
\renewcommand{\arraystretch}{1.1}
\caption{Manipulation Detection Performance Without AEN.
In bold, the best dataset-wise results.}
\label{tab:all_base}
\centering
\footnotesize
\begin{tabular}{R{1.36cm} L{1.68cm} L{1.68cm} L{1.68cm} L{1.68cm} L{1.68cm} L{1.68cm} L{1.68cm} L{1.68cm}}
\toprule
 & \multicolumn{2}{c}{\textbf{MFC18 Dev Dataset}} & \multicolumn{2}{c}{\textbf{MFC18 Eval Dataset}} & \multicolumn{2}{c}{\textbf{Imagenette Dataset}} & \multicolumn{2}{c}{\textbf{UG$^2$ Glider Dataset}} \\
\cmidrule(lr){2-3}
\cmidrule(lr){4-5}
\cmidrule(lr){6-7}
\cmidrule(lr){8-9}
\textbf{Algorithm} & \textbf{GWL1 (\%)} & \textbf{AUC (\%)} & \textbf{GWL1 (\%)} & \textbf{AUC (\%)} & \textbf{GWL1 (\%)} & \textbf{AUC (\%)} & \textbf{GWL1 (\%)} & \textbf{AUC (\%)} \\
\cmidrule(lr){2-3}
\cmidrule(lr){4-5}
\cmidrule(lr){6-7}
\cmidrule(lr){8-9}
\textbf{BLK~\cite{li2009passive}} & 31.23 $\pm$ 10.29 & 59.11 $\pm$ 28.57 & 43.42 $\pm$ 30.75 & 58.12 $\pm$ 25.14 & 31.52 $\pm$ 08.56	& 02.52 $\pm$ 01.22 & 28.87 $\pm$ 07.99	& 0.98 $\pm$ 00.23 \\
\textbf{DCT~\cite{ye2007detecting}} & 35.58 $\pm$ 15.94	& 59.20 $\pm$ 26.80 & 42.98 $\pm$ 36.03 & 54.19 $\pm$ 24.45 & 45.44 $\pm$ 12.13 & 03.84 $\pm$ 04.28 & 48.46 $\pm$ 07.12 & 01.55 $\pm$ 01.71 \\
\textbf{ELA~\cite{ela_krawetz2007}} & 11.95 $\pm$ 09.86	& 56.22 $\pm$ 19.76 & 41.54 $\pm$ 36.58 & 53.93 $\pm$ 16.60 & 15.43 $\pm$ 08.05 & 10.89 $\pm$ 10.70 & 09.68 $\pm$ 02.22 & 03.17 $\pm$ 01.69 \\
\cmidrule(lr){2-3}
\cmidrule(lr){4-5}
\cmidrule(lr){6-7}
\cmidrule(lr){8-9}
\textbf{CFA1~\cite{Dirik2009ImageTD}} & 55.56 $\pm$ 27.77 & 55.54 $\pm$ 23.17& 41.14 $\pm$ 33.50 & 54.88 $\pm$ 22.98 & 57.34 $\pm$ 24.07 & 10.20 $\pm$ 16.16 & 64.43 $\pm$ 25.40 & 08.60 $\pm$ 16.46 \\
\textbf{CFA2~\cite{Ferrara2012ImageFL}} & 28.39 $\pm$ 11.49	& 53.11 $\pm$ 24.88 & 27.05 $\pm$ 25.90 & 56.12 $\pm$ 24.80 & 33.01 $\pm$ 10.11 & 02.91 $\pm$ 02.22 & 36.40 $\pm$ 11.96 & 01.21 $\pm$ 00.68 \\
\cmidrule(lr){2-3}
\cmidrule(lr){4-5}
\cmidrule(lr){6-7}
\cmidrule(lr){8-9}
\textbf{NOI1~\cite{MAHDIAN20091497}} & 15.24 $\pm$ 11.68 & 56.52 $\pm$ 33.81 & 39.43 $\pm$ 36.02 & 54.87 $\pm$ 29.53 & 14.81 $\pm$ 08.25 & 04.14 $\pm$ 04.42 & 13.41 $\pm$ 05.99 & 01.71 $\pm$ 01.46 \\
\textbf{NOI2~\cite{lyu2014exposing}} & \textbf{05.52 $\pm$ 10.50} & 54.37 $\pm$ 18.34 & \textbf{10.86 $\pm$ 17.09} & 52.19 $\pm$ 11.59 & 05.41 $\pm$ 05.18 & \textbf{20.18 $\pm$ 14.99} & \textbf{0.36 $\pm$ 01.25} & \textbf{43.95 $\pm$ 09.41} \\
\textbf{NOI4~\cite{noi4_wagner2015}} & 05.97 $\pm$ 10.27 & 49.30 $\pm$ 17.44 & 15.34 $\pm$ 17.72 & 47.94 $\pm$ 13.61 & \textbf{02.46 $\pm$ 01.75} & 19.94 $\pm$ 08.66 & 01.23 $\pm$ 01.53 & 26.63 $\pm$ 08.40 \\
\cmidrule(lr){2-3}
\cmidrule(lr){4-5}
\cmidrule(lr){6-7}
\cmidrule(lr){8-9}
\textbf{M-Net~\cite{wu2019mantra}} & 10.16 $\pm$ 09.18	& \textbf{62.98 $\pm$ 23.53} & 19.08 $\pm$ 25.96 & \textbf{67.92 $\pm$ 22.40} & 05.39 $\pm$ 02.29 & 06.94 $\pm$ 04.40 & 07.66 $\pm$ 02.63 & 03.03 $\pm$ 01.71 \\
\textbf{RGBN~\cite{zhou2018learning}} & 24.01 $\pm$ 29.35 & 52.67 $\pm$ 16.43 & 24.11 $\pm$	25.54 & 57.61 $\pm$ 17.00 & 56.37 $\pm$	31.41 & 21.82 $\pm$ 15.71 & 87.16 $\pm$	22.70 &	06.42 $\pm$	11.35 \\
\bottomrule
\end{tabular}
\end{table*}



\begin{figure*}[!t]
    \begin{center}
        \includegraphics[width=\textwidth]{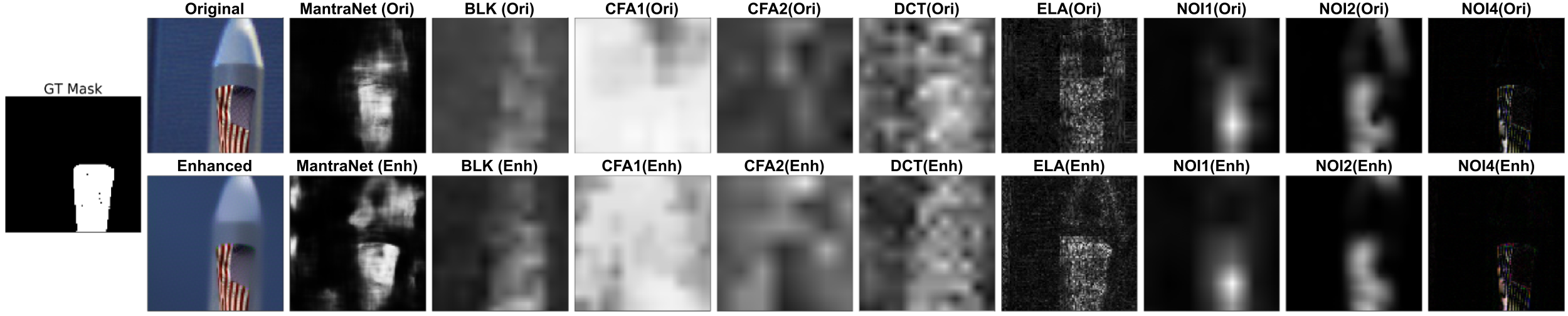}
        \includegraphics[width=\textwidth]{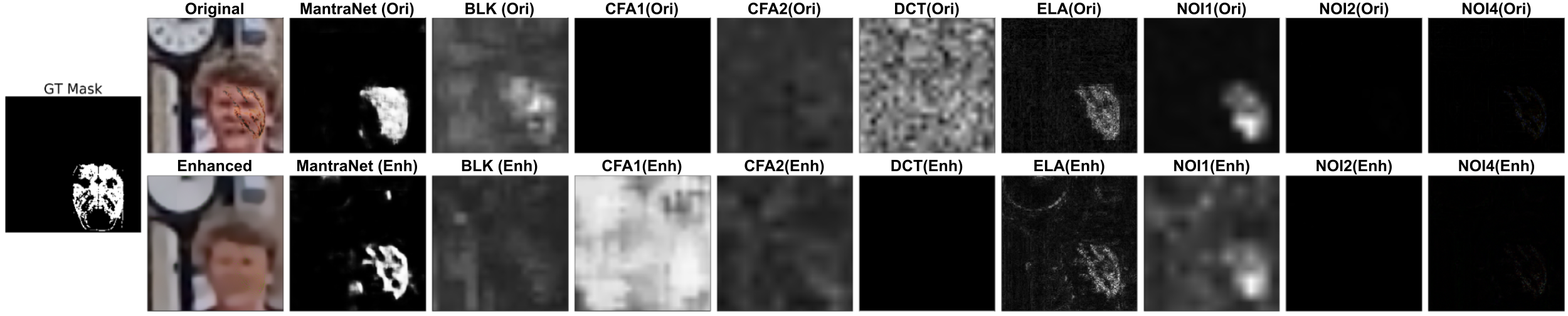}    
    \end{center}
     \caption{Manipulation Detection Results on the AEN Training Set. Nine different algorithms are reported over original and enhanced images from the training dataset, which contains $577$ samples selected from the MFC18 Dev dataset.
    }
    \label{fig:MDevPreds}
\end{figure*}

\begin{figure*}[!t]
    \begin{center}
        \includegraphics[width=\textwidth]{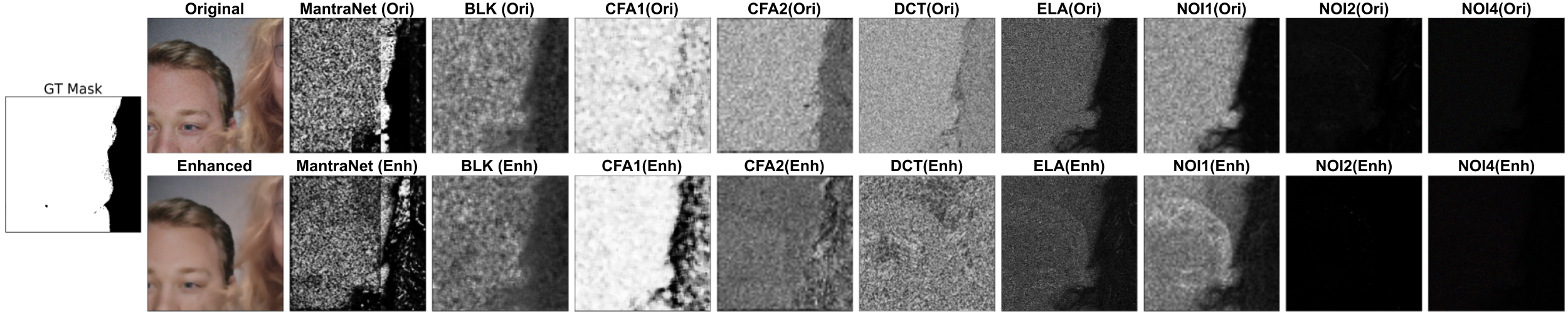} 
        \includegraphics[width=\textwidth]{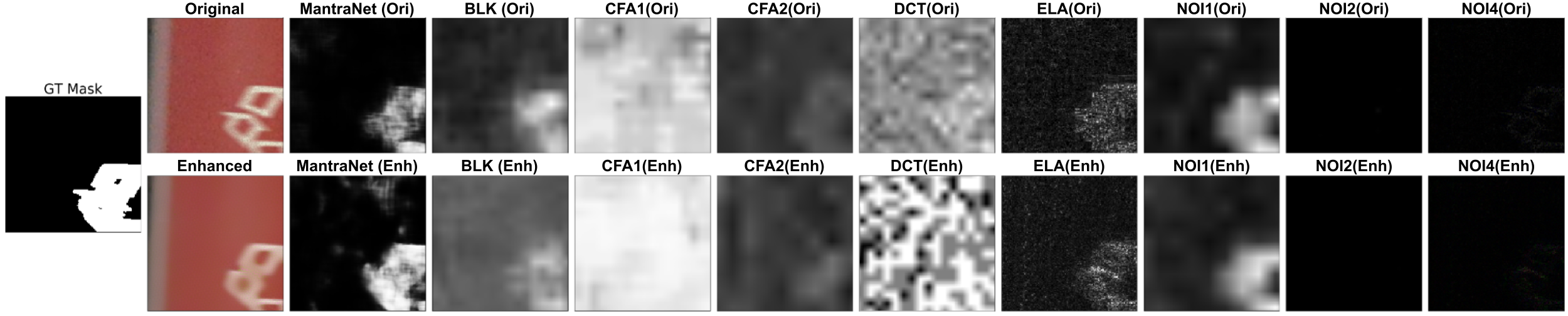} 
    \end{center}
    \caption{
    Manipulation Detection Results on Unseen Data. Nine different algorithms are reported over original and enhanced images, which were not seen by AEN during training time.
    }
    \label{fig:MEvPreds}
\end{figure*}

When looking at the GWL1 metric, noise variance-based approaches outperform other techniques. Out of all the algorithms evaluated, NOI2 consistently produces a low GWL1 score in the predicted detection masks for both datasets. Notably, the noise variance-based approaches tended to generate sparse manipulation masks when compared to the other algorithms, as can be seen in Fig.~\ref{fig:ImgNetPreds}, \ref{fig:MDevPreds}, and~\ref{fig:MEvPreds}. 
This behavior seemed to mimic the sparsity of the ground-truth manipulation masks in the MFC18 data;
on average, the amount of manipulated pixels in the MFC18 Dev and MFC18 Eval datasets is relatively small, with most of the images presenting only $6\%$ and $10\%$ of their pixels manipulated, respectively. 
In comparison, the masks generated by the NOI2 algorithm for the MFC18 Dev dataset presented an average of $8.7\%$ of their pixels detected as manipulated, and while this is close to what we observed from the ground-truth, most of the detected regions had low confidence.
On the other hand, approaches like ELA, NOI1, and Mantra-Net had a high amount of false positives in their gray-scale masks ($67.7\%$, $95\%$, and $94.9\%$ of the pixels were detected as manipulated, respectively).
However, after applying 
different confidence thresholds,
their manipulation detection rates 
become 
more similar to that of the ground-truth ($4.8\%$, $11.2\%$, and $3.8\%$, respectively), which 
explains their better performance 
in terms of AUC.


It is important to note that approaches based on CFA or DCT patterns are particularly sensitive to JPEG compression, and while they might be effective at detecting manipulations on previously uncompressed data, they often struggle with images that have undergone multiple rounds of compression (which is common in images scoured from social media) as this would erode the CFA interpolation discrepancies. This would explain their poor performance on the evaluated data, with CFA1 obtaining the worst performance on both datasets. Despite relying on similar assumptions, CFA2 fares better than its counterpart, which could be a benefit of using a smaller localization window. Having a finer-grained localization allows it to catch any manipulation traces that were not lost after successive rounds of compression. 

In turn, approaches such as ELA rely on the fact that images lose information (particularly high-frequency content) as they are re-compressed after being manipulated. Thus they target the compression residual to find tampered regions. This is especially well-suited for images captured from social media, or images that have undergone multiple rounds of manipulations such as the data present in the MFC18 datasets. Interestingly, ELA demonstrated relatively good performance ($11.95\%$ GWL1) on the MFC18 Dev dataset but struggled with the MFC18 Eval data. 



\subsection{Non-manipulated data}
Most manipulation localization approaches were designed to localize tampered regions in images that are presumed to be doctored, as such, working under the assumption that there is a manipulated region to be found within their inputs. While these approaches aim to minimize the number of false positives within each detection, as we observed from the previous section, this is not always the case, and in fact, some approaches are more prone to detect tampered regions with a low degree of confidence. 

Moreover, the evaluated approaches were designed to identify features within an image that would normally be associated/emphasized by tampering. However, these discrepancies can also be naturally present in pristine data due to the natural capture process of the image, or due to the in-camera processing techniques.
Taking this into account, it is important to measure the performance of manipulation detection methods not only on different types of tampered data but also on pristine samples, in order to get a full understanding of their robustness when applied to real-world scenarios.  To this end, we seek to assess the behavior of the studied localization methods when presented with non-manipulated data. For this, we evaluate their performance on the Imagenette~\cite{imagenette} and UG$^2$ Glider~\cite{8354281} datasets.
We can see in Table~\ref{tab:all_base} (last two column groups) their respective GWL1 and AUC performance.

\begin{figure*}[!t]
    \begin{center}
        \includegraphics[width=\textwidth]{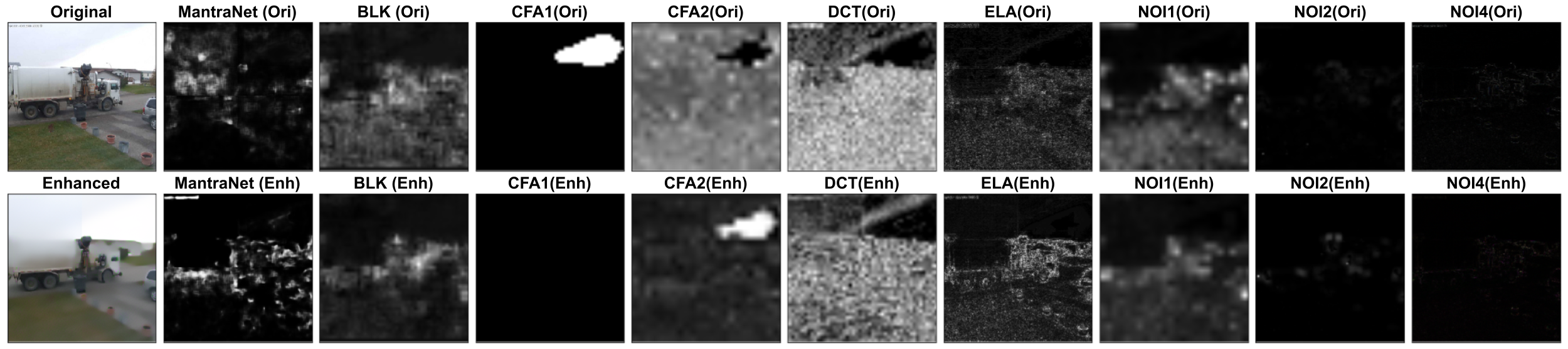}
        \includegraphics[width=\textwidth]{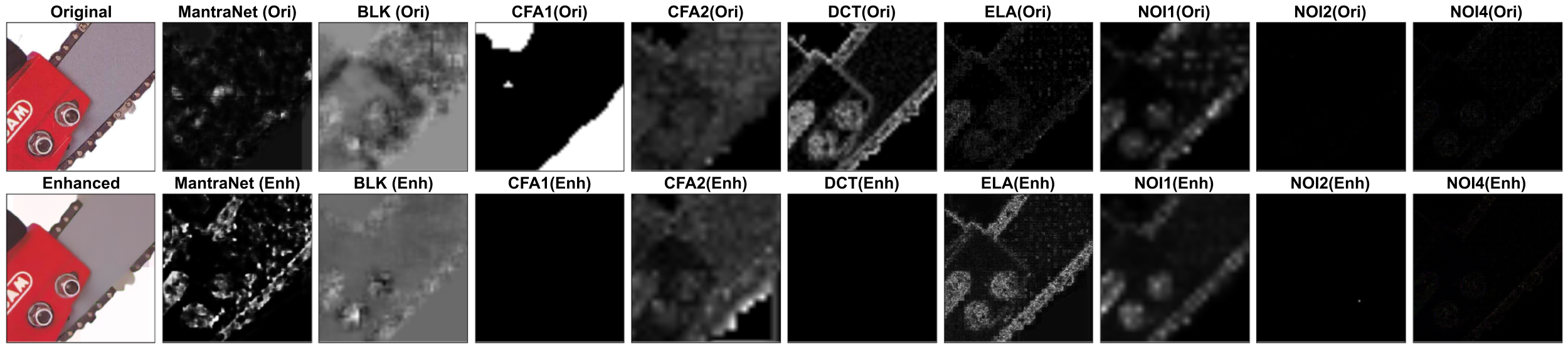}
    \end{center}
    \caption{
    Manipulation Detection Results on the Imagenette Dataset. Nine different algorithms are reported over original and enhanced images.
    Here, the expected manipulation masks are fully black, as there are no maliciously manipulated regions.
    }
    \label{fig:ImgNetPreds}
\end{figure*}

As a subset of ImageNet~\cite{ILSVRC15}, the Imagenette dataset was obtained by scouring the web for images representative of different object classes. As such, some of the images in this dataset might have been subjected to some kind of non-malicious manipulation, such as contrast enhancement, water-marking, or in more extreme scenarios, the removal of the object of interest's background (see the 
third
row of Fig.~\ref{fig:ImgNetPreds}, where a chainsaw has a white background).
Interestingly, the performance of the manipulation detection approaches on this dataset follows similar trends to that on the UG$^2$ Glider Dataset, despite the latter not presenting altered content.

For both datasets, most
of the approaches see a slight reduction (\textit{i.e.}, an improvement) in their gray-scale weighted loss (GWL1) 
when compared to the manipulated MFC18 data. 
In the particular case of the best performing approach (NOI2), the GWL1 error was reduced to $0.36\%$ over the UG$^2$ Glider Dataset.
However, it is interesting to note that this does not translate to their performance when evaluated using AUC, for which there is a dramatic reduction, with most of the algorithms obtaining an AUC under 
$11\%$.
Similarly to what we observed in the manipulated datasets, most of these approaches detect a large number of pixels as manipulated, albeit with significantly low confidence.
This impacts their AUC results, which -- by definition -- take into account their low confidence ranges.
The positive exception relies on the robust NOI2 solution, which presents a significant reduction on its number of false positives.
In the opposite direction, the more recent data-driven learning-based methods suffer from high rates of false positives, with Mantra-Net losing its previous advantage in the face of non-manipulated data.



Notably, the performance of the CFA algorithms was worse when presented with non-manipulated data, especially for the UG$^2$ Glider Dataset. As previously explained, these approaches rely on the presence of discrepancies between the predicted CFA interpolations.
Nonetheless, their prediction of the cameras' CFA patterns might have been disrupted by the video compression present in the dataset, which led to the incongruities between the prediction and the natural pixels being erroneously detected as signs of tampering.

\subsection{Pre-processing Anomaly Enhancement}
As we saw in the previous two subsections, the performance for each of the manipulation detection algorithms varies depending on the type of features they use to determine whether a given region in an image has been tampered with or not. While their ability at detecting manipulation is dependent on the nature of the manipulation, there have been attempts to use deep learning to learn a descriptor that is able to identify multiple types of tampering~\cite{wu2019mantra}. Following this, we seek to assess whether the knowledge of these ``global'' manipulation features can be used to facilitate the detection of the anomalies used by common manipulation detection approaches. This involves the use of a neural network architecture to enhance these features before feeding the ``enhanced'' images to any chosen manipulation detection approach. 

\begin{table*}[!t]
\renewcommand{\arraystretch}{1.1}
\caption{Manipulation Detection Performance With AEN on the MFC18 Datasets.
In bold, results with an improvement over the baseline due to the usage of AEN.
The improvement is noted inside the parentheses.}
\label{tab:mfc18_aen}
\centering
\footnotesize
\begin{tabular}{R{2cm} L{3.3cm} L{3.3cm} L{3.3cm} L{3.3cm}}
\toprule
 & \multicolumn{2}{c}{\textbf{AEN-aided GWL1 (\%)}} & \multicolumn{2}{c}{\textbf{AEN-aided AUC (\%)}} \\
\cmidrule(lr){2-3}
\cmidrule(lr){4-5}
\textbf{Algorithm} & \textbf{MFC18 Dev} & \textbf{MFC18 Eval (Unseen)} & \textbf{MFC18 Dev} & \textbf{MFC18 Eval (Unseen)} \\
\cmidrule(lr){2-3}
\cmidrule(lr){4-5}
\textbf{BLK~\cite{li2009passive}} & \textbf{28.37 $\pm$ 11.52 ($\downarrow$ 02.86)} & \textbf{30.65 $\pm$ 12.45 ($\downarrow$ 01.35)} & \textbf{59.33 $\pm$ 28.40 ($\uparrow$ 00.22)} & \textbf{59.24 $\pm$ 27.70 ($\uparrow$ 01.12)} \\
\textbf{DCT~\cite{ye2007detecting}} & 42.65 $\pm$ 15.60 ($\uparrow$ 07.08) & 43.61 $\pm$ 14.40 ($\uparrow$ 06.66) & 49.12 $\pm$ 19.65 ($\downarrow$ 10.08) & 49.99 $\pm$ 16.05 ($\downarrow$ 04.20) \\
\textbf{ELA~\cite{ela_krawetz2007}} & 13.38 $\pm$ 09.48 ($\uparrow$ 01.43) & 16.66 $\pm$ 14.24 ($\uparrow$ 01.45) & \textbf{57.10 $\pm$ 20.49 ($\uparrow$ 00.88)} & \textbf{54.37 $\pm$ 16.11 ($\uparrow$ 00.44)} \\
\cmidrule(lr){2-3}
\cmidrule(lr){4-5}
\textbf{CFA1~\cite{Dirik2009ImageTD}} & 67.76 $\pm$ 19.14 ($\uparrow$ 12.20) & 65.91 $\pm$ 20.51 ($\uparrow$ 05.98) & \textbf{58.22 $\pm$ 21.85 ($\uparrow$ 02.69)} & \textbf{58.20 $\pm$ 19.66 ($\uparrow$ 03.31)} \\
\textbf{CFA2~\cite{Ferrara2012ImageFL}} & \textbf{27.24 $\pm$ 08.71 ($\downarrow$ 01.15)} & \textbf{29.45 $\pm$ 09.91 ($\downarrow$ 00.31)} & 42.79 $\pm$ 22.21 ($\downarrow$ 10.32) & 51.17 $\pm$ 22.30 ($\downarrow$ 04.95) \\
\cmidrule(lr){2-3}
\cmidrule(lr){4-5}
\textbf{NOI1~\cite{MAHDIAN20091497}} & 16.86 $\pm$ 10.74 ($\uparrow$ 01.63) & 19.18 $\pm$ 13.99 ($\uparrow$ 00.35) & \textbf{59.14 $\pm$ 32.32 ($\uparrow$ 02.62)} & \textbf{59.27 $\pm$ 25.97 ($\uparrow$ 04.39)} \\
\textbf{NOI2~\cite{lyu2014exposing}} & \textbf{05.37 $\pm$ 10.53 ($\downarrow$ 00.14)} & \textbf{09.13 $\pm$ 16.42 ($\downarrow$ 00.17)} & 52.73 $\pm$ 16.64 ($\downarrow$ 01.64) & 51.51 $\pm$ 08.23 ($\downarrow$ 00.69) \\
\textbf{NOI4~\cite{noi4_wagner2015}} & \textbf{05.86 $\pm$ 10.26 ($\downarrow$ 00.10)} & \textbf{09.68 $\pm$ 16.25 ($\downarrow$ 00.16)} & \textbf{52.25 $\pm$ 18.23 ($\uparrow$ 02.95)} & \textbf{51.21 $\pm$ 12.98 ($\uparrow$ 03.27)} \\
\cmidrule(lr){2-3}
\cmidrule(lr){4-5}
\textbf{Mantra-Net~\cite{wu2019mantra}} & 11.73 $\pm$ 08.89 ($\uparrow$ 01.56) & 15.35 $\pm$ 13.99 ($\uparrow$ 01.82) & 61.27 $\pm$ 20.45 ($\downarrow$ 01.71) & 60.50 $\pm$ 20.48 ($\downarrow$ 07.41) \\
\textbf{RGBN~\cite{zhou2018learning}} & 42.63 $\pm$	26.42 ($\uparrow$ 18.62) & 36.49 $\pm$	25.80 ($\uparrow$ 12.38) & \textbf{57.61 $\pm$ 20.97 ($\uparrow$ 04.93)} & \textbf{65.64 $\pm$	20.07 ($\uparrow$ 08.03)} \\
\bottomrule
\end{tabular}
\end{table*}



In light of this, we train the anomaly enhancement network (AEN) described in Section~\ref{sec:aen} and evaluate the detection performance of each of the detection algorithms after pre-processing the data (for both tampered and clean datasets) with it.
As can be seen in Figs.~\ref{fig:ImgNetPreds}, \ref{fig:MDevPreds}, and~\ref{fig:MEvPreds}, the images produced with this new method not only preserve enough of the low-level features to attain similar manipulation masks as their original counterparts, but said manipulation masks tend to have a reduction of low-confidence detections while increasing the confidence on perceived manipulated regions.

Taking into consideration the MFC18 datasets (with manipulated images), the proposed pre-processing step was able to improve the GWL1 loss for four of the algorithms we tested (BLK, CFA2, NOI2, and NOI4), with only major deterioration in the performance for three of the remaining algorithms (DCT, CFA1, and RGBN).
This can be seen through Table~\ref{tab:mfc18_aen} (first and second data columns).



In terms of the AUC 
(see the last two columns of Table~\ref{tab:mfc18_aen}),
the performance for the evaluated detection algorithms on the MFC18 data is ultimately low for both the baseline and enhanced images, starting at $49\%$ and almost reaching only $68\%$. 
Regardless, our method was able to increase the AUC for six of the evaluated approaches (BLK, ELA, CFA1, NOI1, NOI4, and RGBN), with a deterioration in the performance of two of the remaining ones (DCT and CFA2).

\textit{Usage of MFC18 Dev as AEN Training Set:}
The proposed AEN method is data-driven and needs to be trained, as explained in Sec.~\ref{sec:aen}.
To do so, we use $577$ images (out of $757$ available images) from the MFC18 Dev Dataset, which we adopt as the ``AEN Training Set''.
Given we are not able to extract useful patch triplets from all the MFC18 Dev images, $180$ of them remain as unseen samples.
Fig.~\ref{fig:MDevPreds} and~\ref{fig:MEvPreds} provide examples of enhancement on the training set and on unseen images, respectively.
Based on these figures and on 
Table~\ref{tab:mfc18_aen},
we observe similar results when processing both the MFC18 Dev and Eval partitions, suggesting the proposed AEN network translate well to unseen data.



In general, our approach seems to benefit from the short and long skip connections of the 
residual-based
architecture, as it allows 
the preservation and enhancement of 
the noise patterns in the images, which leads to a positive impact on the noise based-methods, 
considering
both GWL1 and AUC metrics. On the other hand, we noticed that even though we did not apply further re-compression to the enhanced images, our method struggled when dealing with JPEG compression-based methods, namely DCT and ELA, with DCT consistently being one of the algorithms most adversely impacted by our technique. This could indicate that while our approach had no issues with preserving the inconsistencies in the blocking artifact grids, it seemed to erode some of the key information that could be used by DCT and ELA to indicate distinct amounts of compression across different image regions. 



As opposed to the MFC18 datasets, we do not have a ground-truth manipulation mask for the data obtained from Imagenette and the UG$^2$ Glider Datasets. Given that this data is presumed to be natural images, our algorithm is expected to apply a minimal set of changes to the image as there would not be any anomalies to be enhanced.
In Fig.~\ref{fig:ImgNetPreds}, we provide qualitative examples of how AEN worked for Imagenette.
In quantitative terms, the AEN-aided GWL1 and AUC values stayed close to the original methods' performance, as we were expecting. 
In the particular case of the Imagenette Dataset, whose images contain enhancements (\textit{e.g.}, non-malicious manipulations such as contrast changes, recoloring, etc.), 
the GWL1 values improved for all but four methods.
Regarding the more challenging Imagenette AUC values, an improvement happened for only three methods, but the AUC deterioration was mostly under one percentage point, except for two solutions (ELA and RGBN).
Focusing on the UG$^2$ Glider Dataset (which does not contain manipulations other than the time watermarks created by the recording devices), AEN improved GWL1 
for all but three methods.
Concerning 
the AUC values, it improved five methods, with a stronger contribution to the learning-based ones (Mantra-Net and RGBN).
All the numbers that give basis to these observations are provided in the Supplementary Material.




\section{Discussion}\label{sec:disc}

 Below we discuss the most important outcomes of this work and answer the three research questions posed in the introduction.

\subsection{Performance on Non-manipulated Data (Question 1)} 
In a scenario where a manipulation detection approach receives as input an image with no discernible local manipulations, it would be expected for it to identify sparse (if any) regions as manipulated with a low degree of confidence. However from our experiments, we have found that this is rarely the case, and in fact, most approaches either identify large areas of the image as tampered with or in fact, point towards well-defined (salient) objects being manipulated with a high degree of confidence in the detection. This highlights the dependence that most of these algorithms have on the assumption that there should be at least two definite classes (manipulated and non-manipulated) to separate the pixels of an image, which could lead to cases where non-doctored images would be mistakenly identified as containing manipulations, with the detectors confidently pointing at arbitrary objects as the culprits.  
Nevertheless, by introducing a pre-processing step designed to enhance any present manipulations in an image, any change in the enhanced image serves the purpose of diminishing noise in the natural image that might otherwise be incorrectly interpreted as a sign of tampering. As such, our method was able to reduce the appearance of low-confidence pixels in the manipulation probability map, which results in cleaner predicted masks. Additionally, the intra-class cohesion property of our approach leads to stronger and more uniform manipulated segments. And while this is a property that greatly facilitates the distinction between the manipulated and non-manipulated regions in a doctored image, it can emphasize naturally occurring anomalies in an unaltered image. Despite this, our method was still able to obtain a localization performance close to the baseline for most of the approaches and even achieved an improvement of over 
$10$ percentage points
for one of the noise-based methods and for the deep-learning approaches we tested.

\subsection{Efficacy of Different Features as Signs of Tampering (Question~2)}
The evaluated detection techniques look into a wide variety of image features to try to identify any tampering. These approaches look at different image patterns present during the image generation process and try to identify any discrepancies in such patterns that might point to the presence of local manipulations, working under the assumption that these discrepancies are more closely related to signs of doctoring than to any naturally occurring distortions. As such, their performance at detecting manipulations of real-life manipulations is strongly tied to the type of features they use to detect tampering as well as the type of manipulation they encounter.

Despite this, we have observed that certain methods are better suited to deal with different types of manipulations across multiple datasets. Namely, we have found that approaches that analyze the noise patterns in the images and use any inconsistencies in them as a sign of tampering, consistently outperform the rest of the techniques, even the deep-learning-based ones. Since the noise patterns in the images remain stable even after multiple rounds of compression, they are well suited for analyzing images from the web that might have been saved and uploaded many times without any tampering applied to them. Moreover, these approaches also benefited from the pre-processing step, provided such a step is able to preserve the noise information in the reconstructed images.

\subsection{Effect of Anomaly Enhancement in Detection (Question 3)} 
Following the idea that local manipulations leave fine traces that can be picked up by localization algorithms, it should be possible for an enhancement method to make such traces easier to perceive not only by the human eye but also by manipulation detection approaches. The experiments conducted in this work showcased the viability of our proposed AEN method as a pre-processing technique to improve the detection of forgeries for a variety of traditional and deep-learning-based manipulation localization algorithms.

The proposed AEN was able to improve the performance of tampering detection approaches based on the analysis of three types of anomalies: (i)~JPEG compression artifacts, (ii)~discrepancies in the CFA patterns, and (iii)~noise variance inconsistencies. While our method was able to either improve or preserve the detection performance for most of the evaluated algorithms, noise-variance-based approaches seemed to benefit the most out of all the evaluated methods, consistently attaining high improvements on both AUC and GWL1. Despite this, our anomaly enhancement is often reflected as a blur-based distortion of the ``anomalous regions'' within an image, which can result in the loss of texture and other high-frequency information. While this can facilitate the recognition of anomalies to the human eyes, this can also distort valuable image statistics and throw off the estimation of the images' CFA interpolation patterns or the grid of blocking artifacts, due to the loss of high-frequency data. In some cases this makes it impossible for the tampering detection methods to provide a correct estimate of the manipulated regions' location.

\begin{figure}[h]
  \centering
    \begin{subfigure}{0.24\textwidth}
      \includegraphics[width=\textwidth]{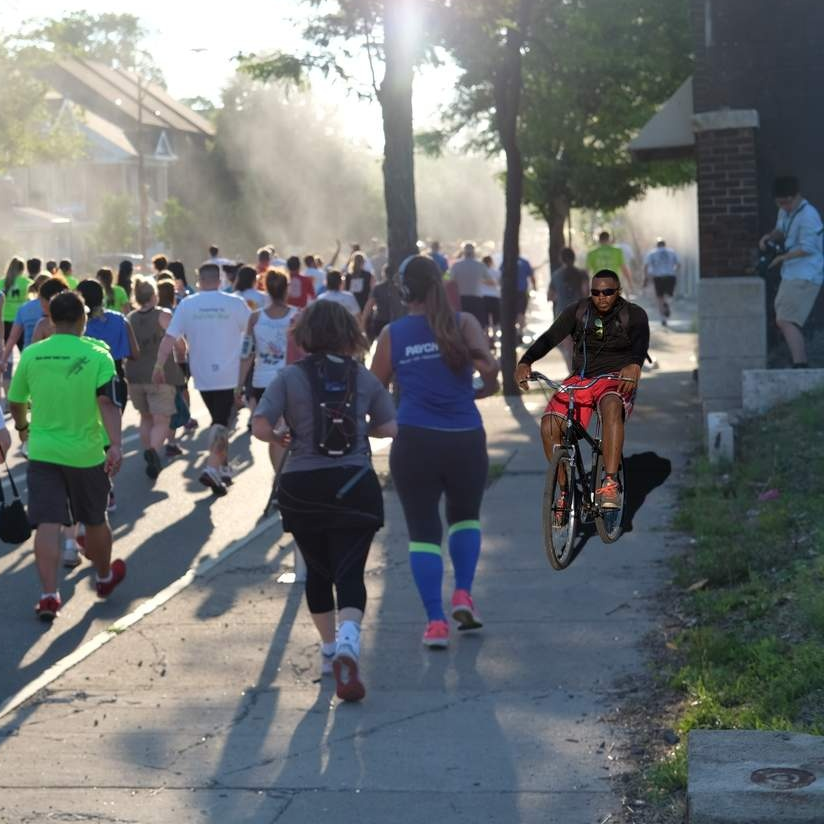}
      \caption{Original}\label{fig:fail_ori}
    \end{subfigure} 
    \begin{subfigure}{0.24\textwidth}
      \includegraphics[width=\textwidth]{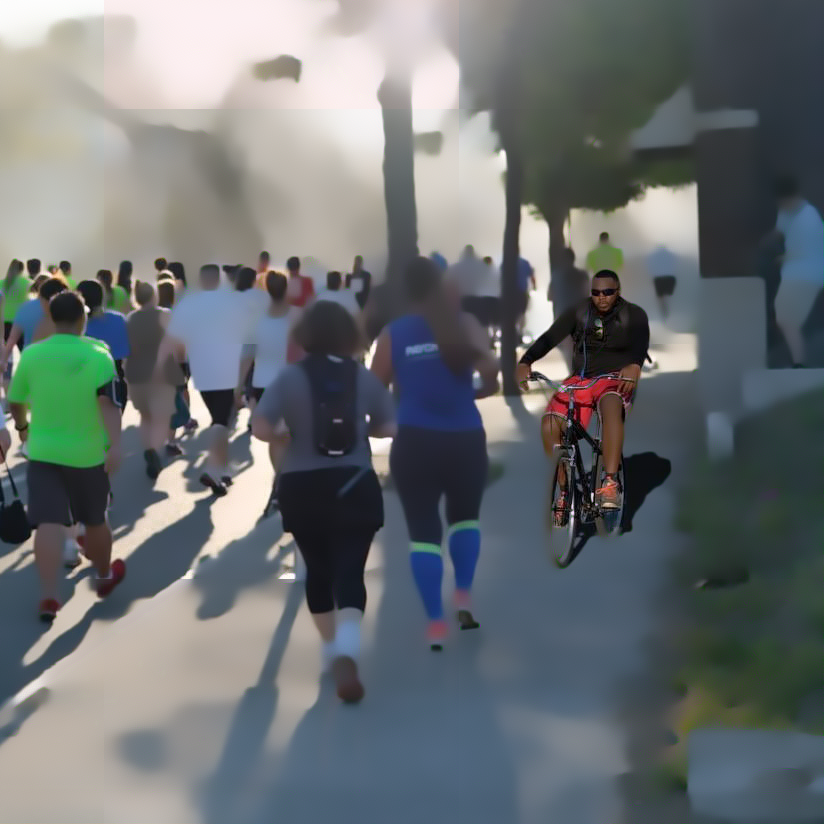}
      \caption{Enhanced}\label{fig:fail_enh}
    \end{subfigure}
  \caption{Failure scenario, where AEN preserves the spliced object and distorts the non-manipulated regions.} \label{fig:fail}
\end{figure}

\subsection{Anomaly Enhancement Limitations}
Our method was able to identify two regions within a tampered image and reconstruct them in such a way that they are visually distinct from one another. This is reflected by one of the regions being reconstructed with a larger amount of distortions.
While one would presume such a region should correspond to the manipulated area in the input image, this is not always the case.  As such there are cases in which the manipulated region is reconstructed faithfully while the authentic regions in the image are distorted (see Fig.~\ref{fig:fail}, for an example). While this effect can single out the boundary between both regions for human observers, when presented to a manipulation detection approach, it results in an inverse (and thus incorrect) prediction mask. 

This might be an effect of our method's learning from disjoint image patches, which might prevent it from identifying a dominant (authentic) region within the image. We thus suggest as future research direction the incorporation of global cues into AEN as a way to address this issue.

\section*{Acknowledgments}
This material is based on research sponsored by the Defense Advanced Research Projects Agency (DARPA) and the Air Force Research Laboratory (AFRL) under agreement number FA8750-20-2-1004. The U.S. Government is authorized to reproduce and distribute reprints for Governmental purposes notwithstanding any copyright notation thereon. The views and conclusions contained herein are those of the authors and should not be interpreted as necessarily representing the official policies or endorsements, either expressed or implied, of DARPA and AFRL or the U.S. Government.


{\small
\bibliographystyle{IEEEtran}
\bibliography{refs}
}


\begin{IEEEbiography}[{\includegraphics[width=1in,height=1.25in,clip,keepaspectratio]{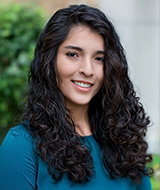}}]{Rosaura G. VidalMata} is a Ph.D.~Candidate in the Computer Science and Engineering Department at the University of Notre Dame. She received her B.S.~degree in Computer Science at the Tecnologico de Monterrey (ITESM) in 2015, where she graduated with an Honorable Mention of Excellence. Her research interests include computer vision, machine learning, and biometrics.
\end{IEEEbiography}\vspace{-3em}

\begin{IEEEbiography}[{\includegraphics[width=1in,height=1.25in,clip,keepaspectratio]{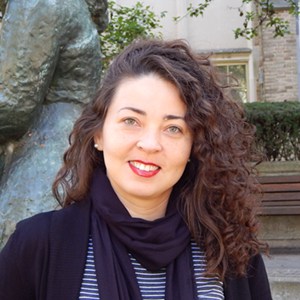}}]{Priscila Saboia} is a Data Engineer with the Center for Research Computing at the University of Notre Dame, USA. She received her B.S. degree at the Federal University of Par\'{a} (UFPA), Brazil, and her M.S. degree at the State University of Campinas (Unicamp), Brazil, both in computer science. Her research interests include data science, scientific computing, and machine learning.
\end{IEEEbiography}\vspace{-3em}

\begin{IEEEbiography}[{\includegraphics[width=1in,height=1.25in,clip,keepaspectratio]{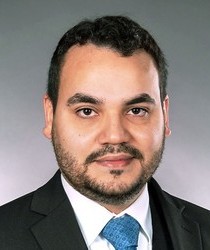}}]{Daniel Moreira} received a B.S.~degree from the Federal University of Par\'{a}, Brazil, in 2006, an M.S.~degree from the Federal University of Pernambuco, Brazil, in 2008, and the Ph.D.~degree from the University of Campinas, Brazil, in 2016, all in computer science. After working for four years as a Systems Analyst with the Brazilian Federal Data Processing Service (SERPRO), he joined the University of Notre Dame for six years, first as a Post-Doctoral Fellow and later as an Assistant Research Professor. He is currently an Assistant Professor in the Department of Computer Science at Loyola University Chicago. His research interests include media forensics, computer vision, machine learning, and biometrics.
\end{IEEEbiography}\vspace{-3em}

\begin{IEEEbiography}[{\includegraphics[width=1in,height=1.25in,clip,keepaspectratio]{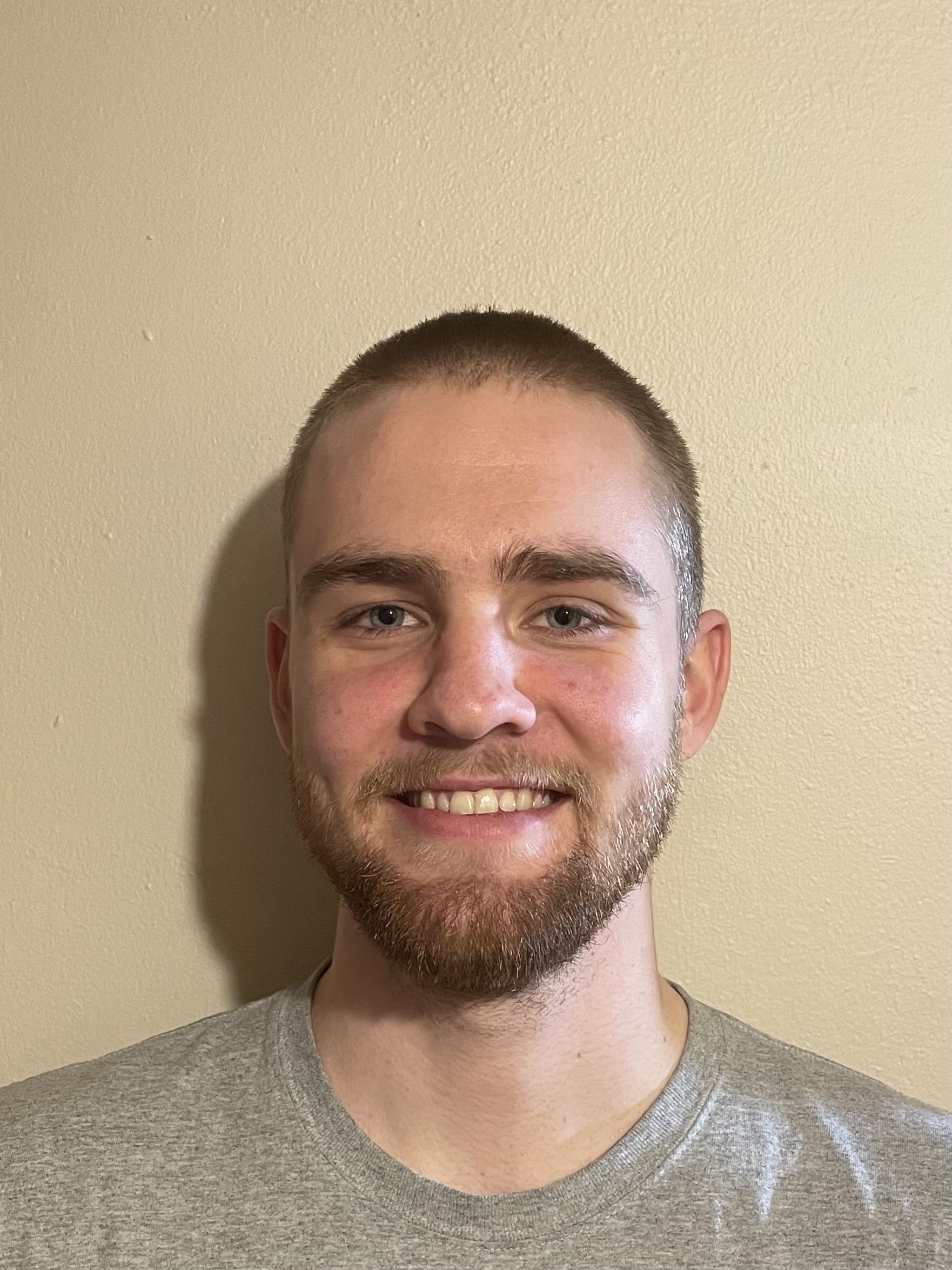}}]{Grant Jensen} is an undergraduate student at UW-Madison studying math and computer science. His interests include creating end-to-end machine learning pipelines and promoting open-source machine learning solutions.
\end{IEEEbiography}\vspace{-3em}

\begin{IEEEbiography}[{\includegraphics[width=1in,height=1.25in,clip,keepaspectratio]{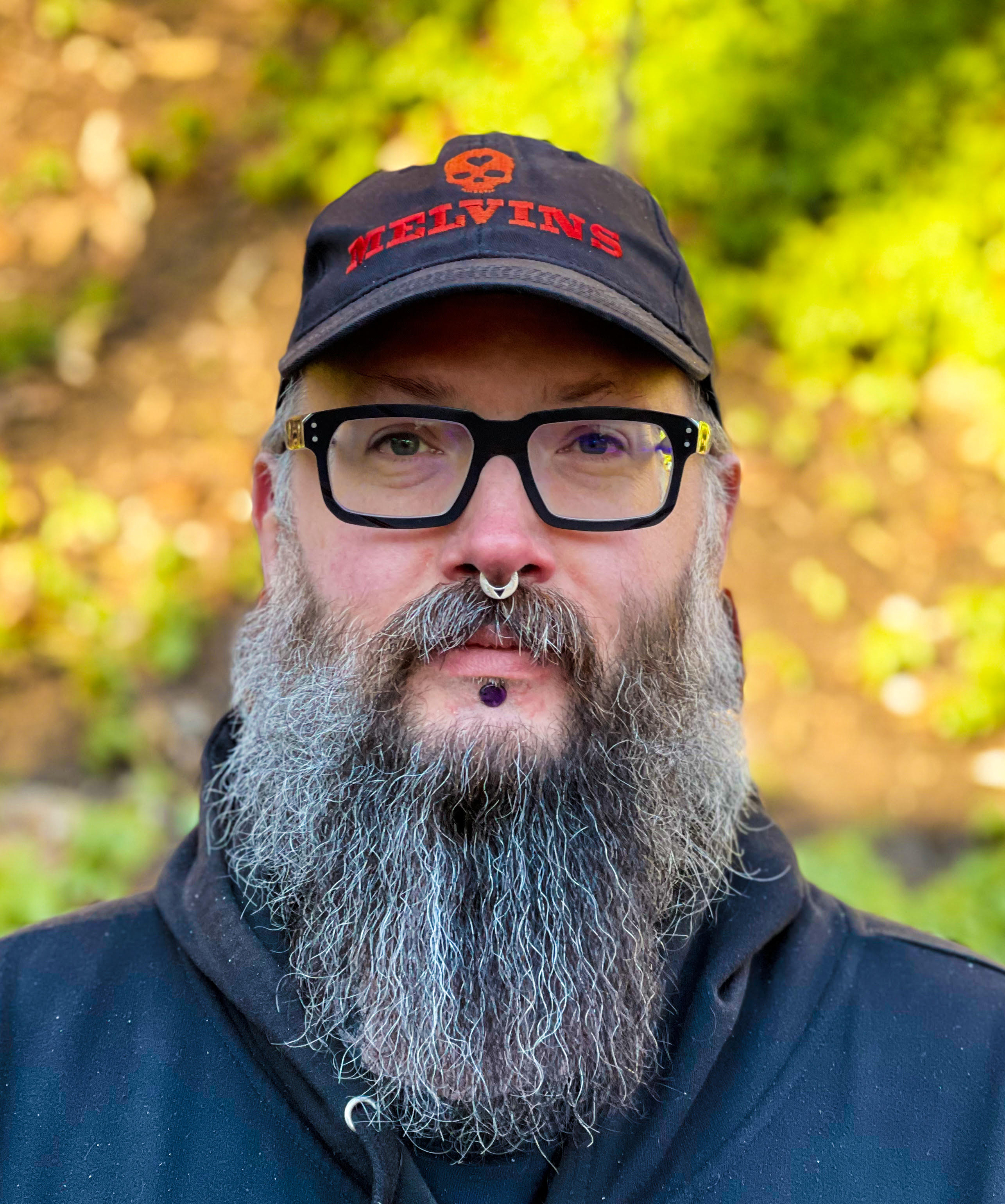}}]{Jason Schlessman}is Principal Software Engineer at Red Hat Research, focused on novel artificial intelligence and machine learning innovations that lead to pragmatic and feasible solutions. He especially targets projects that serve the well-being of humanity, fostering ethical uses of technology. He can be found online @ EldritchJS.
\end{IEEEbiography}\vspace{-3em}

\begin{IEEEbiography}[{\includegraphics[width=1in,height=1.25in,clip,keepaspectratio]{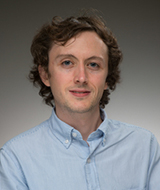}}]{Walter J. Scheirer} received the M.S. degree in computer science from Lehigh University, in 2006, and the Ph.D. degree in engineering from the University of Colorado, Colorado Springs, CO, USA, in 2009. He is an Associate Professor with the Department of Computer Science and Engineering, University of Notre Dame. Prior to joining the University of Notre Dame, he was a Postdoctoral Fellow with Harvard University from 2012 to 2015, and the Director of Research and Development with Securics, Inc., from 2007 to 2012. His research interests include computer vision, machine learning, biometrics, and digital humanities. 
\end{IEEEbiography}\vspace{-3em}



\end{document}


\title{Supplementary Material of \\\textit{On the Effectiveness of Image Manipulation Detection in the Age of Social Media}}
\author{}

\maketitle

This document supplements the content of the manuscript entitled ``On the Effectiveness of Image Manipulation Detection in the Age of Social Media,'' by providing more details of the experimental setup and obtained results.

\section{Evaluation Metrics}
We evaluate the performance of each of the discussed algorithms following the metrics defined by the MFC18 Challenge~\cite{mfc18_challenge}.
All of the following metrics were computed through the Mediscore tool~\cite{mediscore_nist_2021}, which was provided by the authors of the dataset. 

Most metrics depend on having binarized values for the predicted manipulation masks, whereas the localization algorithms more often than not generate a gray-scale localization mask, where the value of each pixel ranges from 0 to 255.
While it is possible to binarize the predicted masks by either specifying a binarization threshold $\theta$, the selection of such threshold then impacts the performance and sensitivity of such metrics.
Taking this into account, we use two metrics that do not depend on binary masks: Area Under the Curve (AUC) and Grayscale Weighted L1 Loss (GWL1).

\subsection{Area Under the Curve}
Given the possibility of having multiple thresholds, we can use the area under the Receiver Operating Characteristic (ROC) curve as a measure of the system's performance across a wide range of threshold values.
This metric is able to indicate how much the model is capable of distinguishing between two classes.
The higher the AUC the better the model is at distinguishing between each class. 

Therefore, for a given ground-truth mask $m$, and a predicted mask $\hat{m}$, we calculate the relation between their true positive rate (TPR) and false positive rate (FPR) after binarizing the predicted mask using a range of $N$ thresholds $\theta_{n} \in \{\theta_{0}, ..., \theta_{N-1}\}$:
\begin{dmath}
    AUC(m, \hat{m}) = 
    \hfill \frac{1}{2} \sum_{\theta_{n}}  \left [ TPR (m, \hat{m}_{\theta_{n+1}}) +  TPR (m, \hat{m}_{\theta_{n}}) \right ] 
    \hfill \times \left [ FPR (m, \hat{m}_{\theta_{n+1}}) -  FPR (m, \hat{m}_{\theta_{n}})\right ].
\end{dmath} 

\subsection{Gray-scale Weighted L1 Loss}
Each of the evaluated methods outputs a gray-scale manipulation mask, where the value for each pixel is correspondent to the confidence of that pixel being manipulated. 
As such, we also report the Weighted L1 Loss for the gray-scale (GWL1) masks.
Contrary to the AUC, the lower the GWL1 value, the better the solution.

To make the evaluation of the masks more robust to the complexity of their borders, the Mediscore tool defines a no-score region around each ground-truth mask $m$.
This region is defined as:
\begin{equation}
    NoScore = Dilation(m) - Erosion(m). 
\end{equation}
Any pixels within the no-score region are then ignored for scoring purposes.

The GWL1 for a predicted mask $\hat{m}$ in the face of a ground-truth mask $m$ is then defined as:
\begin{equation}
\label{eq_wl1}
GWL1(m, \hat{m}) = \frac{1}{ER} \sum_{i=1}^{P} w_{i} {\frac{\left | m_{i} - \hat{m}_{i} \right |}{255}},
\end{equation}
where $m_{i}$ is the $i$-th pixel in the mask $m$, $P = size(m) = size(\hat{m})$ is the number of pixels within each mask, and $ER = size(MR) + size(NotMR)$ is the number of pixels of the Evaluated Regions $MR$ and $NotMR$.
In turn, $MR = Erosion(m)$ is the region scored as the correct manipulated region, and $NotMR = m - Dilation(m)$ is the region scored as the correct non-manipulated region.
Finally, the weight $w_{i}$ used for each pixel within the evaluated mask is:
\begin{equation}
    w_{i} = \begin{Bmatrix}
    0, & \mathrm{if} \; i\: \in Dilation(m) \; \mathrm{and} \; i \not\in Erosion(m)\\ 
    1, & \mathrm{otherwise} \hfill
    \end{Bmatrix}. 
\end{equation}

Since the Mediscore tool does not support the analysis of non-manipulated images, when evaluating the performance of non-manipulated data (Imagenette and UG$^2$ datasets), we mark a small arbitrary region (10x10 pixels) in each image as manipulated and use it as ground truth.
This allows us to still use Mediscore to compare and rank the different solutions when they are presented with non-manipulated images, at the cost of a small penalty for the 10x10 regions, equally applied to all the images and methods. 
Given the small area of this fictional tampering, any outputs passing as true positives have little impact on the final loss score.

\begin{table*}[!t]
\renewcommand{\arraystretch}{1.1}
\caption{Manipulation Detection Performance With AEN on the Imagenette and UG$^2$ Glider Datasets.
In bold, results with an improvement over the baseline due to the usage of AEN.
The improvement is noted inside the parentheses.}
\label{tab:non_man_aen}
\centering
\footnotesize
\begin{tabular}{R{2cm} L{3.3cm} L{3.3cm} L{3.3cm} L{3.3cm}}
\toprule
 & \multicolumn{2}{c}{\textbf{Imagenette Dataset}} & \multicolumn{2}{c}{\textbf{UG$^2$ Glider Dataset}} \\
\cmidrule(lr){2-3}
\cmidrule(lr){4-5}
\textbf{Algorithm} & \textbf{AEN-aided GWL1 (\%)} & \textbf{AEN-aided AUC (\%)} & \textbf{AEN-aided GWL1 (\%)} & \textbf{AEN-aided AUC (\%)} \\
\cmidrule(lr){2-3}
\cmidrule(lr){4-5}
\textbf{BLK~\cite{li2009passive}} & \textbf{29.02 $\pm$ 09.24 ($\downarrow$ 02.50)} & \textbf{02.52 $\pm$ 01.22 ($\uparrow$ 00.00)} & \textbf{24.45 $\pm$ 07.87 ($\downarrow$ 04.41)} & \textbf{00.98 $\pm$ 00.23 ($\uparrow$ 1.13E-05)} \\
\textbf{DCT~\cite{ye2007detecting}} & 46.33 $\pm$ 10.48 ($\uparrow$ 00.89) & 03.31 $\pm$ 03.73 ($\downarrow$ 00.53) & \textbf{44.17 $\pm$ 11.33 ($\downarrow$ 04.29)} & \textbf{03.76 $\pm$ 10.34 ($\uparrow$ 02.21)}  \\
\textbf{ELA~\cite{ela_krawetz2007}} & 16.70 $\pm$ 07.63 ($\uparrow$ 01.27) & 05.13 $\pm$ 04.18 ($\downarrow$ 05.76) & 10.82 $\pm$ 05.03 ($\uparrow$ 01.13) & 02.61 $\pm$ 01.01 ($\downarrow$ 00.56) \\
\cmidrule(lr){2-3}
\cmidrule(lr){4-5}
\textbf{CFA1~\cite{Dirik2009ImageTD}} & 59.08 $\pm$ 23.69 ($\uparrow$ 01.74) & 09.26 $\pm$ 15.59 ($\downarrow$ 00.94) & 76.03 $\pm$ 10.49 ($\uparrow$ 11.60) & 01.77 $\pm$ 06.06 ($\downarrow$ 06.83) \\
\textbf{CFA2~\cite{Ferrara2012ImageFL}} & \textbf{31.02 $\pm$ 09.24 ($\downarrow$ 01.99)} & 02.83 $\pm$ 02.02 ($\downarrow$ 00.08) & \textbf{28.48 $\pm$ 08.56 ($\downarrow$ 07.92)} & 01.05 $\pm$ 00.41 ($\downarrow$ 00.16) \\
\cmidrule(lr){2-3}
\cmidrule(lr){4-5}
\textbf{NOI1~\cite{MAHDIAN20091497}} & \textbf{13.65 $\pm$ 07.42 ($\downarrow$ 01.16)} & 03.51 $\pm$ 03.49 ($\downarrow$ 00.63) & 14.78 $\pm$ 06.53 ($\uparrow$ 01.37) & 01.11 $\pm$ 00.50 ($\downarrow$ 00.61) \\
\textbf{NOI2~\cite{lyu2014exposing}} & \textbf{04.73 $\pm$ 05.03 ($\downarrow$ 00.68)} & \textbf{24.53 $\pm$ 14.16 ($\uparrow$ 04.35)} & \textbf{00.18 $\pm$ 00.50 ($\downarrow$ 00.18)} & \textbf{45.45 $\pm$ 06.52 ($\uparrow$ 01.50)} \\
\textbf{NOI4~\cite{noi4_wagner2015}} & \textbf{02.24 $\pm$ 01.66 ($\downarrow$ 00.22)} & 19.40 $\pm$ 08.76 ($\downarrow$ 00.54) & \textbf{00.85 $\pm$ 00.83 ($\downarrow$ 00.38)} & 24.21 $\pm$ 08.44 ($\downarrow$ 02.42) \\
\cmidrule(lr){2-3}
\cmidrule(lr){4-5}
\textbf{Mantra-Net~\cite{wu2019mantra}} & 08.02 $\pm$ 03.42 ($\uparrow$ 02.63) & \textbf{16.80 $\pm$ 07.52 ($\uparrow$ 09.86)} & \textbf{06.40 $\pm$ 01.34 ($\downarrow$ 01.26)} & \textbf{10.45 $\pm$ 05.06 ($\uparrow$ 07.42)} \\
\textbf{RGBN~\cite{zhou2018learning}} & \textbf{56.37 $\pm$ 31.41 ($\downarrow$ 04.02)} & 21.82 $\pm$ 15.71 ($\downarrow$ 02.01) & \textbf{75.67 $\pm$ 26.22	 ($\downarrow$ 11.48)} & \textbf{12.16 $\pm$	13.11 ($\uparrow$ 05.74)} \\
\bottomrule
\end{tabular}
\end{table*}

\section{AEN Results on the Imagenette and UG$^2$ Glider Datasets}
In Table~\ref{tab:non_man_aen}, we report the impact of the proposed Anomaly Enhancement Network (AEN) as a pre-processing step over the Imagenette and UG$^2$ Glider Datasets, considering both the GWL1 and AUC metrics.
Here, it is important to highlight that these datasets do not present maliciously manipulated content.
As a consequence, we do not have ground-truth manipulation masks for the images that compose them.

While the UG$^2$ Glider Dataset does not contain manipulations other than the time watermarks created by the recording devices, the Imagenette Dataset does include a number of different global manipulations, such as watermarks, contrast enhancement, and recoloring.
Despite the effect these global manipulations might have on the features used by the manipulation detectors, our enhancement approach was able to maintain 
the Imagenette GWL1 values, which stayed close to the original data's.
Only four methods had an increase in GWL1 (all under 3 percentage points).
Regarding the Imagenette AUC, our method improved the performance of 3 solutions. However, for the remaining manipulation detection approaches, the performance decrease was mostly under 1 percentage point, 
with the exception of ELA and RGBN.

Focusing on the UG$^2$ Glider Dataset (see the last two columns of Table~\ref{tab:non_man_aen}), 
AEN improved GWL1 for all but three methods.
About the more challenging AUC metric, it improved for five methods, with a stronger contribution to the learning-based ones (Mantra-Net and RGBN).

\section{Impact of Different Augmentation Goals}

\begin{figure}[!t]
\centering
\includegraphics[width=0.48\textwidth]{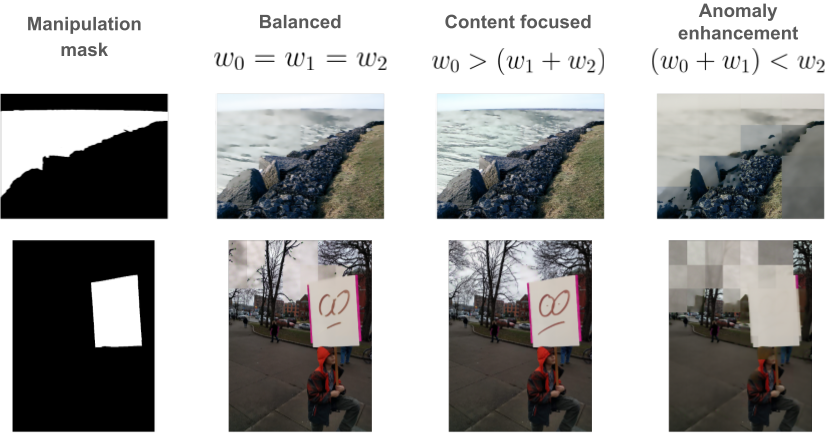}
\caption{Visualization of the effects of the different learning goals on a manipulated image.}
\label{fig:learningGoal}
\end{figure}

Our learning objective focuses on three main goals, (1) content restoration, (2) anomaly enhancement, and (3) intra-class cohesion. Our results section showcases the performance of a network trained with a balanced focus on all three objectives ($w_{1} = w_{2} = w_{3}$, as per Eq.~\ref{eq:loss}). 

\begin{equation}
\begin{split}
L(a, \hat{a}, p, n) = w_{0} D(a, \hat{a}) + w_{1} D(f(\hat{a}), f(p)) \\- w_{2} D(f(\hat{a}), f(n)).
\end{split}\label{eq:loss}
\end{equation}

We observed that giving a larger weight to the content restoration term of our loss produces a higher quality reconstruction (see Fig.~\ref{fig:learningGoal}) with a slightly lower performance than when using a balanced loss (an average of $0.2$-percentage-point difference for both GWL1 and AUC). Given the premise of our approach is to increase the discrepancies between anomalous and original regions, we also tested making this element of our learning the main focus during training. Interestingly while the qualitative results show a sharper contrast between manipulated and non-manipulated regions, in some cases this learning goal increased the distance between the original and enhanced features, but not the distance between the manipulated and non-manipulated data, this was reflected by a sharp decrease in the performance on AUC (with an average decrease of $2.2$ percentage points).

{\small
\bibliographystyle{IEEEtran}
\bibliography{sm-refs}
}